\documentclass{article}

\usepackage{PRIMEarxiv}

\usepackage[utf8]{inputenc} 
\usepackage[T1]{fontenc}    
\usepackage{hyperref}       
\usepackage{url}            
\usepackage{booktabs}       
\usepackage{amsfonts}       
\usepackage{nicefrac}       
\usepackage{microtype}      
\usepackage{lipsum}
\usepackage{fancyhdr}       
\usepackage{graphicx}       
\graphicspath{{media/}}     

\usepackage{times}
\usepackage{soul}
\usepackage{amsmath}
\usepackage{amsthm}
\usepackage{booktabs}
\usepackage{algorithm}
\usepackage{algorithmic}
\usepackage{enumerate}
\usepackage{subfigure}
\usepackage{multirow}
\usepackage{enumitem}
\usepackage{xspace}
\usepackage{multicol}
\usepackage{diagbox}
\newtheorem{definition}{Definition}

\newcommand{\mname}{$\texttt{C}^{\texttt{3}}\texttt{TS}$\xspace}

\title{\mname: \texttt{C}onfidence-guided Learning Process for \texttt{C}ontinuous \texttt{C}lassification of \texttt{T}ime \texttt{S}eries
\thanks{\textit{\underline{Citation}}: 
\textbf{Chenxi Sun, Moxian Song, Derun Cai, Baofeng Zhang, Shenda Hong, and Hongyan Li. 2022. Confidence-Guided Learning Process for Continuous Classification of Time Series. In Proceedings of the 31st ACM International Conference on Information and Knowledge Management (CIKM ’22), Octo- ber 17–21, 2022, Atlanta, GA, USA. ACM, New York, NY, USA, 5 pages. https://doi.org/10.1145/3511808.3557565}} 
}

\author{
  Chenxi Sun \\
  Key Laboratory of Machine Perception\\
  (Ministry of Education), Peking University\\
  Beijing, China\\
  School of Artificial Intelligence,\\ Peking University\\
  Beijing, China\\
  \texttt{sun\_chenxi@pku.edu.cn} \\
   \And
  Moxian Song \\
  Key Laboratory of Machine Perception\\
  (Ministry of Education), Peking University\\
  Beijing, China\\
  School of Artificial Intelligence,\\ Peking University\\
  Beijing, China\\
  \texttt{songmoxiani@pku.edu.cn} \\
    \And
  Derun Cai \\
  Key Laboratory of Machine Perception\\
  (Ministry of Education), Peking University\\
  Beijing, China\\
  School of Artificial Intelligence,\\ Peking University\\
  Beijing, China\\
  \texttt{cdr@stu.pku.edu.cn} \\
    \And
  Baofeng Zhang \\
  Key Laboratory of Machine Perception\\
  (Ministry of Education), Peking University\\
  Beijing, China\\
  School of Artificial Intelligence,\\ Peking University\\
  Beijing, China\\
  \texttt{boffinzhang@stu.pku.edu.cn} \\
    \And
  Shenda Hong* \\
  National Institute of Health Data Science,\\Peking University\\
  Beijing, China\\
  Institute of Medical Technology, \\Health Science Center of Peking University\\
  Beijing, China\\
  \texttt{hongshenda@pku.edu.cn} \\
    \And
  Hongyan Li* \\
  Key Laboratory of Machine Perception\\
  (Ministry of Education), Peking University\\
  Beijing, China\\
  School of Artificial Intelligence,\\ Peking University\\
  Beijing, China\\
  \texttt{leehy@pku.edu.cn} \\
}

\begin{document}
\maketitle

\begin{abstract}
In the real world, the class of a time series is usually labeled at the final time, but many applications require to classify time series at every time. e.g. the outcome of a critical patient is only determined at the end, but he should be diagnosed at all times to facilitate timely treatment. Thus, in this paper, we propose a new concept: Continuous Classification of Time Series (CCTS). There are two open problems about CCTS: (1) Data arrangement. Time series is a kind of dynamic data. It evolves multiple distributions over time. The division of multi-distribution will directly affect the classification accuracy; (2) Model training strategy. When a model learns multi-distributed data, it always forgets the old distribution or overfits in the main distribution. Different data learning orders will result in different model performances. We find that the process of model learning multiple distributions can be similar to the process of human learning multiple knowledge. Thus, we propose a novel Confidence-guided method for CCTS to arrange the data and schedule the training, named \mname. It imitates the objective-confidence and the alternating self-confidence of humans in their learning process, which is described by the Dunning-Kruger Effect. Specifically, we define an importance-based objective-confidence to arrange and replay data, and design an uncertainty-based self-confidence to control the training duration. Experimental results on four real-world datasets show that our method is more accurate than all baselines at every time.
\end{abstract}


\section{Introduction}

\begin{figure*}[t]
\centerline{
\begin{minipage}[t]{0.38\linewidth}
\centering
\includegraphics[width=\linewidth]{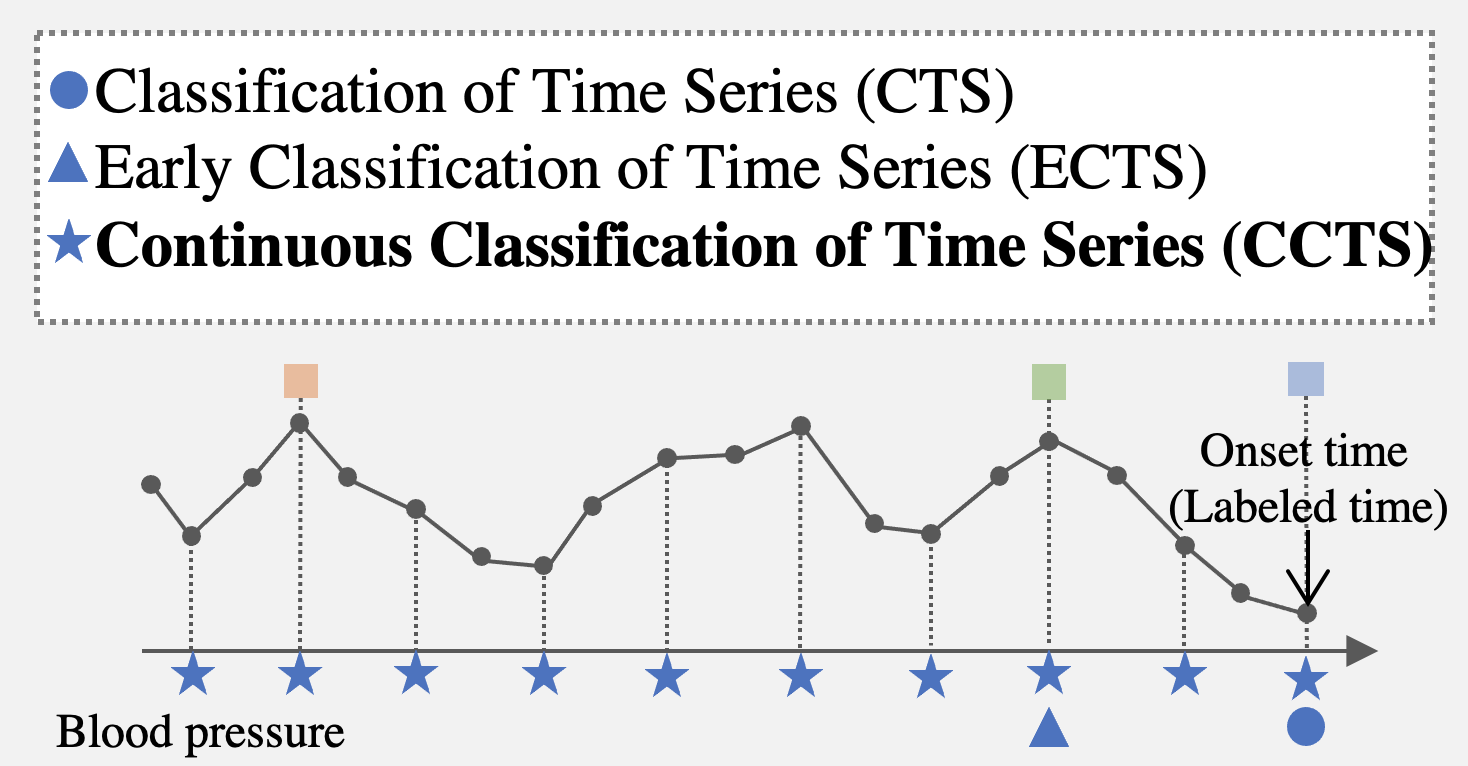}
\end{minipage} 
\begin{minipage}[t]{0.38\linewidth}
\centering
\includegraphics[width=\linewidth]{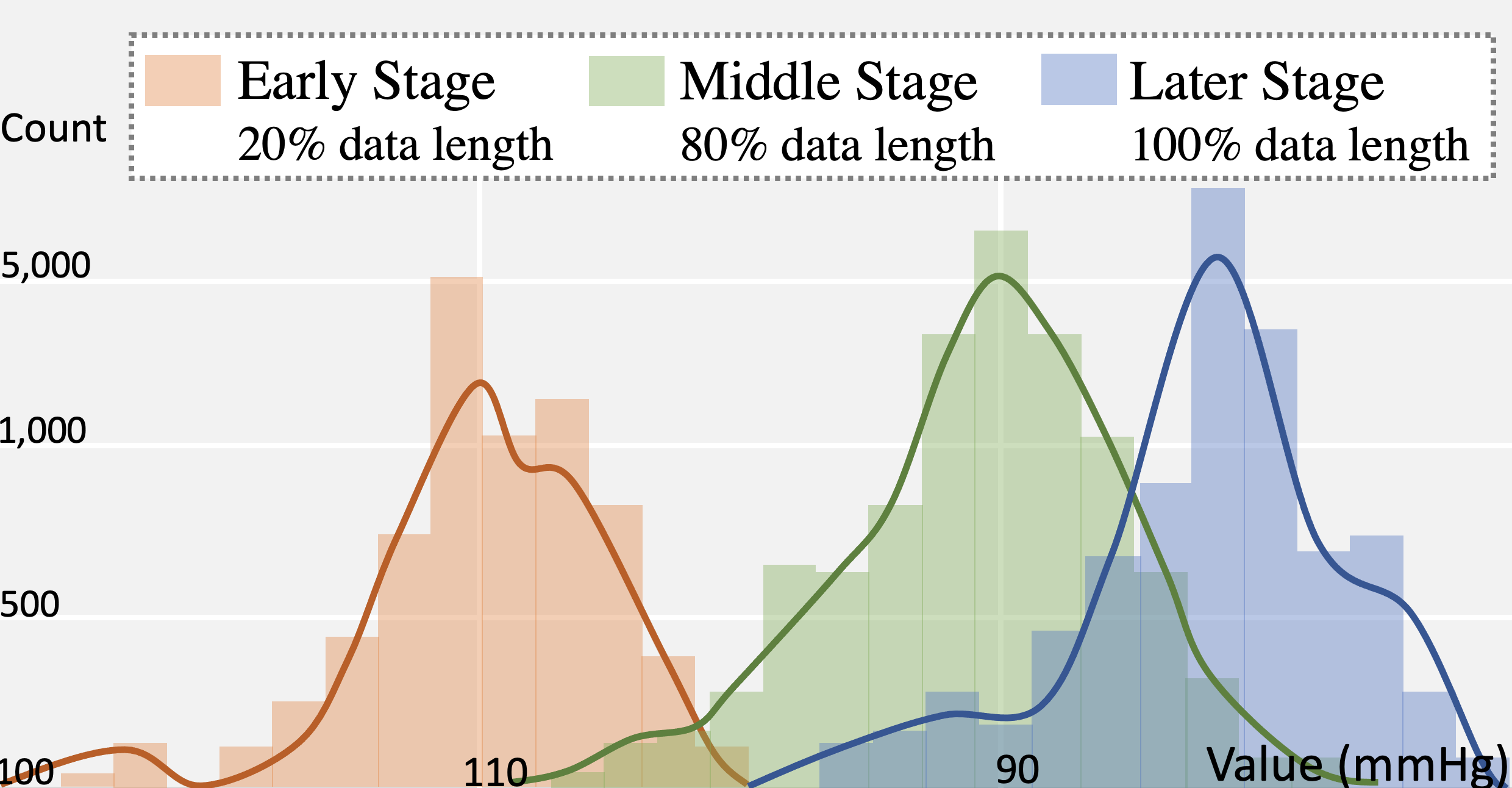}
\end{minipage}
\begin{minipage}[t]{0.38\linewidth}
\centering
\includegraphics[width=\linewidth]{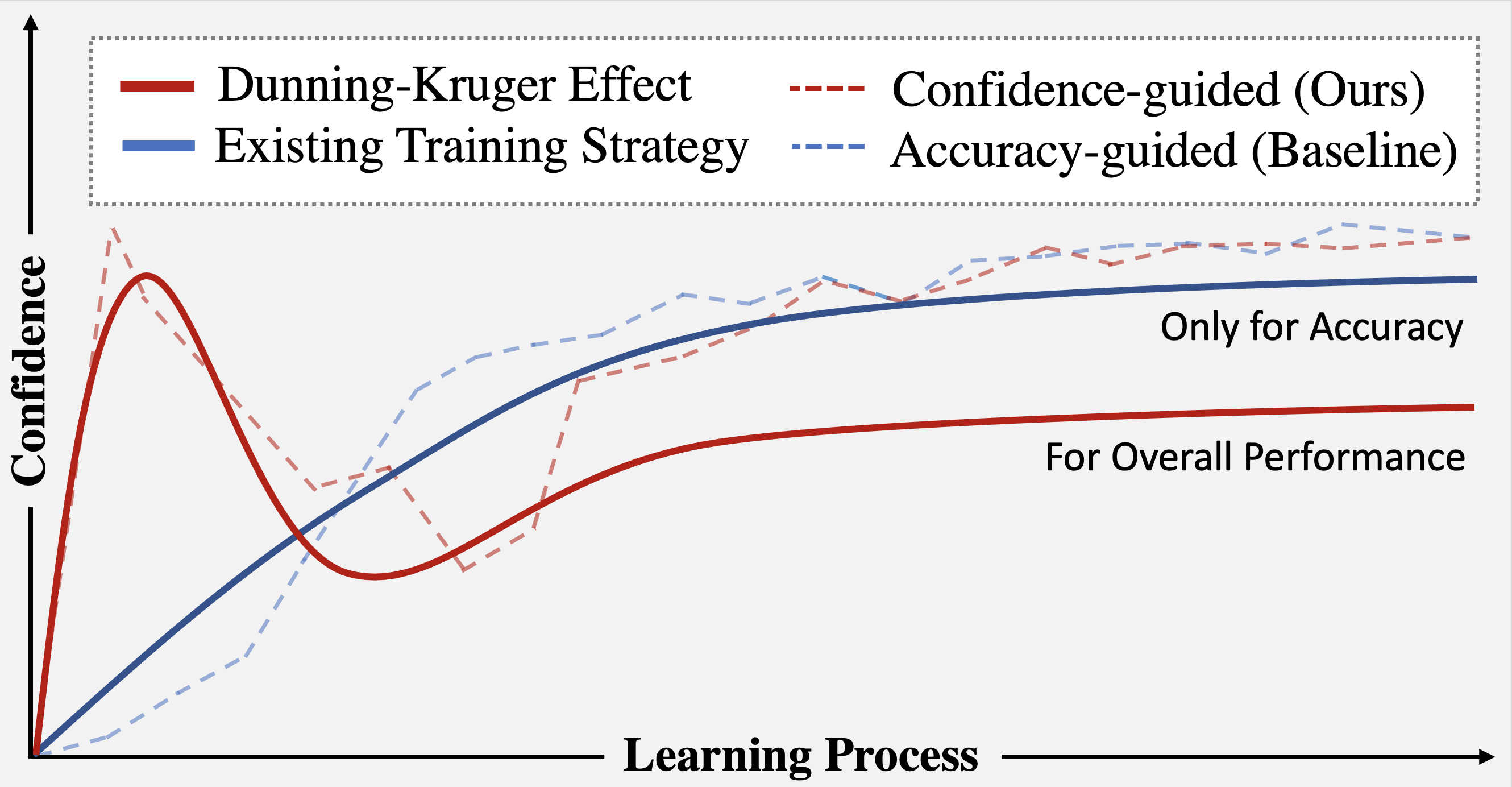}
\end{minipage}}
\caption{Continuous Classification of Time Series (CCTS) \& Multi-distribution in Time Series \& Confidence Change during Learning\\
\footnotesize{The left figure shows that different from existing classification tasks of time series (CTS and ECTS), CCTS needs the continuous classification mode; The middle figure shows that the statistics \protect\cite{2021A} of blood pressure of 2,000 sepsis patients represent multiple distributions among early, middle and later stages; The right figure shows the confidence changes of human with an alternating process: ignorance$\rightarrow$overconfidence$\rightarrow$disappointment$\rightarrow$development.}} \label{fig:introduction}
\end{figure*}

In the real world, many applications need to classify time series at every time. For example, in the Intensive Care Unit (ICU), most detected vital signs change dynamically with the evolution of deceases. The status perception is needed at any time as the real-time diagnosis provides more opportunities for doctors to rescue lives \cite{2014The}. But patient labels, e.g., mortality and morbidity, are unknown in early stages but available only at the onset time. In response to the current demand, we propose a new concept -- Continuous Classification of Time Series (CCTS), to classify time series at every early time before the final labeled time, as shown in Figure \ref{fig:introduction} left.

CCTS task requires the method to learn data from different advanced stages. But for most practical time series, changed data characteristics lead to the evolved data distribution, and finally produce the data form of multi-distribution. As shown in Figure \ref{fig:introduction} middle, data distributions of blood pressure of sepsis patients vary among early, middle, and later stages, bringing a triple-distribution. Thus, to achieve CCTS, the method should model such multi-distributed data. 
And there are two open problems:

\textbf{ How to prepare multi-distribution before training the model?} As CCTS requires the mode of continuous classification, all subsequences of a time series had better be learned. The intuition is to get multi-distribution according to time stages. But the optimal time interval is unclear: small time stages produce overlapping distributions while large time stages produce distinct distributions. They all affect model learning by worsening the forgetting or overfitting \cite{DBLP:journals/nn/ParisiKPKW19}. Some methods define distributions by the data complexity \cite{DBLP:conf/acl/DaiLLSHSZ20}, e.g., the number of objects \cite{DBLP:journals/pami/WeiLCSCFZY17} or rare words \cite{DBLP:conf/acl/XuZMWXZ20}. But compared with the data form of image and text, time series is more abstract, the complexity is hard to tell. Besides, data that is difficult for people may not be difficult for machines. e.g., human-identifiable stable vital signs are more likely to confuse the model \cite{DBLP:journals/tai/GuptaGBD20}.

\textbf{ How to train the model under multi-distribution?} A single model, like deep neural network, is lack of ability to learn all distributions simultaneously \cite{9349197} as they are restricted by the premise of independent identically distributed (i.i.d) data \cite{DBLP:conf/aaai/ShimMJSKJ21}. As shown in Figure \ref{fig:introduction3}, frequent learning of new knowledge will inevitably lead to the forgetting of old ones \cite{DBLP:journals/nn/ParisiKPKW19}, and too much training on one distribution may make the parameter fall into the local solution, resulting in poor generalization \cite{DBLP:conf/iclr/SahaG021}. Besides, a model will receive different parameter matrices according to different training orders \cite{DBLP:journals/corr/abs-2010-13166}. As shown in Figure \ref{fig:introduction4}, training a model from early time series to late and from late time series to early will lead to different result accuracy and convergence speed. Thus, feeding examples in a meaningful order but randomly is critical \cite{DBLP:journals/corr/abs-2010-13166}.

To solve both two problems, we propose a novel Confidence\footnote{\footnotesize{The confidence in this paper represents the human cognition of their ability, different from the confidence in Statistics, the probability of the result in the confidence interval.}}-guided method for CCTS, named \mname. As we assume that the process of model learning multiple distributions can be similar to the process of human learning multiple knowledge: people will first arrange the learning order (data arrangement), then control the progress of learning and review according to their mastery (model learning). The mastery of knowledge is usually assessed by human confidence, they will relearn the knowledge that they are not confident about and improve its priority in the learning order. This human behavior is the Dunning-Kruger effect \cite{2000Unskilled,DBLP:conf/acl/ZhouYWWC20} as shown in Figure \ref{fig:introduction} right. It is an alternating process with experiences of ignorance, overconfidence, disappointment and development:

When a human learns a new field of knowledge, he will first scratch and grasp the overall framework and therefore has a lot of confidence. Then, he begins to find that he really had little in-depth knowledge and loses his confidence. Over time, he studies deeply, becoming more and more experienced and confident \cite{2021Dunning}. 

To imitate human confidence-guided learning, we define the confidence from two aspects of objective-confidence and self-confidence. Specifically, we design the importance-based objective-confidence by using the importance coefficient. It arranges data and makes the model relearn important samples to consolidate memory loosely; We design the uncertainty-based self-objective by defining the total uncertainty. It controls the training duration under the current distribution and schedules the training order among distributions. Experimental results on four real-world datasets show that \mname is more accurate than all baselines at every time.

\section{Related Work}\label{sec:related work}

The popularity of time series classification has attracted increasing attention in many practical fields \cite{DBLP:conf/iknow/SantosK16}. The existing work can be summarized into two categories. And more detailed work and concept distinction are in Appendix.

\subsection{ One-shot classification: classifying at a fixed time}

The foundation is Classification of Time Series (CTS), classifying the full-length time series \cite{DBLP:journals/datamine/FawazFWIM19}. But in time-sensitive applications, Early Classification of Time Series (ECTS) is more critical, making classification at an early time \cite{DBLP:journals/tai/GuptaGBD20}. For example, early diagnosis helps for sepsis outcomes \cite{DBLP:conf/aaai/LiuLSGN18}. However, both CTS and ECTS give the one-shot classification, only classifying once and just lean a single data distribution. They have good performances on i.i.d data at a fixed time, like early 6 hours sepsis diagnosis \cite{2019Early}, but fail for multi-distribution. In fact, continuous classification is composed of multiple one-shot classifications as shown in Figure \ref{fig:introduction} left.

\subsection{Continuous classification: classifying at every time}

To achieve CCTS, learning multi-distribution is essential. Most methods use multi-model to model multi-distribution, like SR \cite{DBLP:journals/tnn/MoriMDL18} and ECEC \cite{lv2019effective}. They divide data according to time stages and design different classifiers for different distributions. But they only focus on the data division with no analysis or experiment to show the rationality. Besides, the operation of classifier selection in multi-model will result in additional losses.

A model will face the catastrophic forgetting problem when learning multi-distributed data. Recently, continual learning \cite{9349197} methods aim to address the issue of static models incapable of adapting their behavior for new knowledge. It learns a new task at every moment and each task corresponds to one data distribution. Replay-based methods \cite{2021A,DBLP:conf/aaai/IseleC18} retrain the model with old data to consolidate memory and focus on the storage limitation. But in their settings, the old task and new task are clear so that the multi-distribution is fixed. This is different from the scenario of CCTS, where the distributions are not determined and need to be divided and defined firstly. Besides, one continual learning contains multiple tasks, their data distributions are not particularly similar. But in CCTS, it aims at one task, thus the replay easily causes the over-fitting problem. 

Training a model by multiple distributions should also consider the training order. A subdiscipline named curriculum learning \cite{DBLP:conf/icml/BengioLCW09}, where an instinctive strategy presents an easy-to-hard order, referring to human courses from simple to difficult. Many methods define the difficulty through the complexity of data, e.g., the number of object in an image \cite{DBLP:journals/pami/WeiLCSCFZY17}, the number of rare words in a sentence \cite{DBLP:conf/acl/XuZMWXZ20}. But compared with the data form of image and text, time series is more abstract, and the complexity is hard to define. Some features, such as Shaplets \cite{DBLP:journals/isci/LiangW21} and frequency \cite{DBLP:conf/aaai/LaiXLZ15}, merely consider one aspect of time series which fails to fully cope with the data difficulty for a model. Besides, the difficulty of human definition may not match the machine. For example, stable vital signs are more likely to confuse the model and lead to classification errors \cite{DBLP:journals/tai/GuptaGBD20}.

The existing human behavior-imitated methods assume that the confidence monotonically increases in the training process \cite{DBLP:conf/isocc/JoP20}, which does not conform to the Dunning-Kruger effect \cite{2000Unskilled,DBLP:conf/acl/ZhouYWWC20}. As ignorance more frequently begets confidence than does knowledge\footnote{\footnotesize{Quote from Charles Robert Darwin (1809-1882)}}, the increasing confidence may not lead to growing knowledge.

The uncertainty \cite{DBLP:conf/nips/KendallG17} is similar to the confidence \cite{DBLP:conf/acl/QuirkLD18}. It can measure if the model parameters could describe the data distribution \cite{DBLP:conf/aaai/XiaoW19}. The uncertainty can be quantified by using the variance of probabilistic distribution of model parameters \cite{DBLP:conf/acl/ZhouYWWC20}. Some researchers design more general methods, using dropout mechanism \cite{DBLP:conf/icml/GalG16} and prior networks \cite{DBLP:conf/nips/MalininG18} to assist in the evaluation. Some curriculum learning works use data uncertainty to arrange the initial training order, like data competence \cite{DBLP:journals/corr/abs-1811-00739,DBLP:conf/naacl/PlataniosSNPM19} and joint difficulty \cite{DBLP:conf/acl/ZhouYWWC20}. In CCTS, in addition to the uncertainty, we need to combine the feature of time series data form, avoid forgetting problem to design the model confidence, and consider both distribution order and learning duration.

\begin{figure*}[t]
\centering
\includegraphics[width=0.95\linewidth]{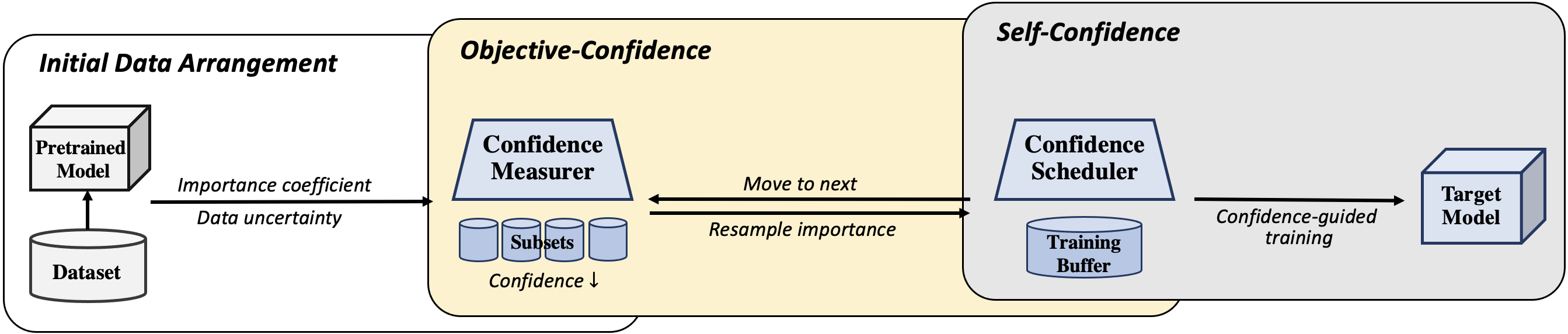}
\caption{Confidence-Guided Training Process for Continuous Classification of Time Series (\mname)}
\label{fig:method}
\end{figure*}

\section{Problem Formulation}

Without the loss of generality, we use the univariate time series to present the problem. Multivariate time series can be described by changing $x_{t}$ to $x_{t}^{i}$. $i$ is the i-th dimension. 

\begin{definition} [Continuous Classification, CC] \label{def:CC}
A time series $X=\{x_{1},...x_{T}\}$ is labeled with a class $C\in \mathcal{C}$ at the final time $T$. CC classifies $X$ at every $t$ with loss $\sum_{t=1}^{T}\mathcal{L}(f(X_{1:t}),C)$.
\end{definition}

Note that unlike continuous classification, CTS and ECTS are just one-shot classification, where they optimize the objective with a single loss $\mathcal{L}(f(x),c)$. Our CCTS should consider the multi-distribution and classify more times. Thus, we define CCTS as:

\begin{definition} [Continuous Classification of Time Series CCTS]\label{def:CCTS} A dataset $\mathcal{S}$ contains $N$ time series. Each time series $X=\{x_{1},...x_{T}\}$ is labeled with a class $C\in \mathcal{C}$ at the final time $T$. As time series varies among time, it has a subsequence series with $M$ different distributions $\mathcal{D}=\{\mathcal{D}^{1},...,\mathcal{D}^{M}\}$, each $\mathcal{D}^{m}$ has subsequence $X_{1:t^{m}}$. CCTS learns every $\mathcal{D}^{m}$ and introduces a task sequence $\mathcal{M}=\{\mathcal{M}^{1},...,\mathcal{M}^{M}\}$ to minimize the additive risk $\sum_{m=1}^{M}\mathbb{E}_{\mathcal{M}^{m}}[\mathcal{L} (f^{m}(\mathcal{D}^{m};\theta),C)]$ with model $f$ and parameter $\theta$. $f^{m}$ is the model $f$ after being trained for $\mathcal{M}^{m}$. When the model is trained for $\mathcal{M}^{m}$, its performance on all observed data cannot degrade: $\frac{1}{m}\sum_{i=1}^{m}\mathcal{L}(f^{i},\mathcal{M}^{i}) \leq \frac{1}{m-1}\sum_{i=1}^{m-1} \mathcal{L}(f^{i},\mathcal{M}^{i})$.
\end{definition}

\section{Confidence-guided Method}

We introduce the confidence from objective-confidence and self-confidence. They are achieved by the importance coefficient and the uncertainty during the model training process.

\subsection{Objective-Confidence ----- Replay by Importance Coefficient} \label{sec:Importance Coefficient}

People gain the objective-confidence through regular examinations or tests, where they can know their weak knowledge through test results and learn it again. Thus, we use the classification accuracy as the test results of the model and apply the importance coefficient to find the weak knowledge of the model and replay it again to improve the objective-confidence of the model.

We focus on the adaptive method to explore a wider space, where the replayed data is dynamic and determined according to the current state. We introduce an importance-based replay method. In each round, it only re-trained the model by some important samples. The importance of each sample is learned from the importance coefficient in the objective of an additive loss function. 

The model learns the importance of a time series $X_{i}$ to it by the coefficient $\beta_{i}$ of $X_{i}$'s loss $\mathcal{L}_{i}$. The overall loss is the sum of the loss of each sample in current distribution $\mathcal{D}$:
\begin{equation}\label{eq:importance}
\mathcal{L}=-\frac{1}{|\mathcal{D}|}\sum_{X_{i}\in\mathcal{D}}\beta_{i}^{2}(-\mathcal{L}_{i})+\lambda(\beta_{i}-1)^{2}
\end{equation}

The importance coefficient $\beta_{i}$ of each sample $X_{i}$ is learned/updated by the gradient descent $\beta_{i}\leftarrow \beta_{i}-\frac{\partial\mathcal{L}}{\partial \beta{i}}$. If a sample $X_{i}$ is hard to classify, its loss $\mathcal{L}_{i}$ will be larger and $-\mathcal{L}_{i}$ will be smaller. In order to minimize the overall loss, its $\beta_{i}$ will be larger. Based on this, the important samples for the model are those difficult to learn with $\beta>\epsilon$. Meanwhile, as $\beta$ is the coefficient of loss, if $\beta=0$, the loss are hard to be optimized. Thus, inspired by \cite{2021A}, we introduce a regularization term $(\beta-1)^{2}$ and initialize $\beta=1$ to penalize it when rapidly decaying toward $0$.

\subsection{Self-Confidence ----- Approximate by Uncertainty Evaluation}

People gain the current self-confidence through their abilities and the problem difficulty. Correspondingly, we approximate the self-confidence of the model by defining the total uncertainty according to the model uncertainty and the data uncertainty. 

The high self-confidence of the model shows that the model believes it already has good classification ability, which means the uncertainty is low. A small score of total uncertainty indicates the model is confident that the current training data has been well learned, and the termination of the decline in scores represents the signal to shift to the next training stage.
\begin{equation} \label{eq:model confidence}
Confidence=\frac{1}{U_{total}}
\end{equation}

The existing work shows that the total uncertainty = epistemic uncertainty + aleatoric uncertainty \cite{DBLP:conf/nips/MalininG18}:
\begin{equation} \label{eq:unceratinty}
\small
    \underbrace{\mathcal{H}[\mathbb{E}_{P(\theta|\mathcal{D})}[P(y|x^{*},\theta)]]}_{Total}=\underbrace{\mathcal{I}[y,\theta|x^{*},\mathcal{D}]}_{Epistemic}+\underbrace{\mathbb{E}_{P(\theta|\mathcal{D})}[\mathcal{H}[P(y|x^{*},\theta)]]}_{Aleatoric}
    \nonumber
\end{equation} 

Inspired by this idea, we define the total uncertainty as:
\begin{equation} 
U_{total}=U_{model}+U_{data}
\end{equation}

First, we define the model uncertainty as the variance $\sigma$ of a distribution of $K$ classification probabilities for data $X$. The classification results are predicted by the model with $K$ different parameter disturbances. For reasons of computational efficiency, we adopt widely used Monte Carlo Dropout \cite{DBLP:conf/icml/GalG16} to approximate Bayesian inference and achieve the parameter disturbances. For the current distribution $\mathcal{D}^{m}=\{X_{1:t^{m}},C\}$, the model uncertainty of model $f(\theta)$ is:
\begin{equation} \label{eq:model unceratinty}
    U_{model}(f(\theta),\mathcal{D}^{m})=\frac{1}{|\mathcal{D}^{m}|}\sum_{(x_{1:t^{m}},c)\in\mathcal{D}^{m}}\sigma_{k=1}^{K} (p(c|x_{1:t^{m}},\theta_{k}))
\end{equation}

Second, we define a novel importance-aware data uncertainty function. It is based on the entropy of the predictive distribution \cite{DBLP:conf/nips/Lakshminarayanan17}. It behaves similar to max probability, but represents the uncertainty encapsulated in the entire distribution. We also inject the learned importance coefficient $\beta$ of the data $x$ to data uncertainty. Thus, the most uncertain data for the model is that has the largest combination value of entropy and importance. The uncertainty of a data $x$ is:
\begin{equation} \label{eq:data unceratinty}
    U_{data}(x)=\frac{\beta}{|\mathcal{C}|}\sum_{c\in \mathcal{C}}p(c|x)log(1-p(c|x))
\end{equation}

Finally, according to the model uncertainty and the data uncertainty, we can get the total uncertainty for the current model $f^{m}$:
\begin{equation} \label{eq:total unceratinty}
\begin{aligned}
U_{total}(f^{m})=&U_{model}(f^{m},\mathcal{D}^{m})+U_{data}(\mathcal{D}^{m})\\
=&\frac{1}{|\mathcal{C}|}\sum_{c\in{\mathcal{C},x\in\mathcal{D}^{m}}}U_{model}(f^{m},(x,c))\\
&+\frac{1}{|\mathcal{D}^{m}|}\sum_{x\in\mathcal{D}^{m}}U_{data}(x)\\
\end{aligned}
\end{equation}

\begin{algorithm}[t]\caption{\mname} \label{alg:C3TS}
\begin{algorithmic}[1] 
\REQUIRE 
Training set $\mathcal{S}=\{(X^{i},C^{i})\}_{i=1}^{N}$;\\
\quad  \ \ An untrained target model $f^{1}$;\\
\quad  \ \ An untrained same model as the target model $f'$.
\ENSURE 
A well-trained target model $f^{M}$.
\STATE // \textsc{Data Arrangement (Pre-train)}
\STATE Extend $\mathcal{S}\leftarrow\{(X_{1:t},C)\}_{t=1}^{T}$
\STATE Train $f'$ by $\mathcal{S}$ using loss in Equation \ref{eq:importance}
\STATE Get $U_{data}$ for each $X\in \mathcal{S}$ of $f'$ using Equation \ref{eq:data unceratinty}
\STATE Split $\mathcal{S}$ into $M$ datasets $\mathcal{D}=\{\mathcal{D}^{m}\}_{n=1}^{M}$ by $U_{data}$ and define $M$ baby steps $\mathcal{M}=\{\mathcal{M}^{m}\}_{m=1}^{M}$ by $\mathcal{D}$ with increasing $U_{data}$\\
\STATE //\textsc{Objective-confidence (Task scheduling)}
\STATE Initialize current training set buffer $\mathcal{B}=\mathcal{D}^{1}$
\FOR {$m = 1$ to $M$}
    \STATE // \textsc{Self-confidence (Duration scheduling)}
    \WHILE{not early stop of Equation \ref{eq:model confidence}}
    \STATE Train $f^{m}$ by $\mathcal{B}$ using loss $U_{total}$ in Equation \ref{eq:total unceratinty}
    \ENDWHILE
    \STATE Get $\beta$ for each $X\in \mathcal{B}$ by $f^{m}$ using Equation \ref{eq:importance}
        \STATE Update $\mathcal{B}\leftarrow\{\mathcal{D}^{m+1},\{X_{i}|\beta_{i}>\epsilon\}\}$
\ENDFOR
\end{algorithmic}
\end{algorithm}

\subsection{Confidence-Guided Training Process}

\mname consists of three interrelated and cooperative modules: Initial data arrangement module, objective-confidence scheduling module, and self-confidence scheduling module.

\begin{itemize}[leftmargin=10 pt]
    \item The \textit{initial data arrangement module} gives an initial data learning order for the model based on confidence. It imitates the fact that before starting a study, students will arrange the learning order according to the knowledge difficulty.
\end{itemize}

It is a pre-train process as shown in the white part of Figure \ref{fig:method}. We first expand the original data set $\mathcal{S}$ by extracting $T$ subsequences $\{X_{1:t}\}_{t=1}^{T}$ from each time series $X=\{x_{1},...,x_{T}\}$. The subsequence $X_{1:t}$ is an early $T-t$ time data. We train the model from early $T-1$ time dataset to early $0$ time dataset with loss in Equation \ref{eq:importance}. Then, we can get the importance coefficient $\beta$ of each sample in $\mathcal{S}$. The initial data learning order is based on the model's confidence in data, which is calculated by the importance-aware data uncertainty in Equation \ref{eq:data unceratinty}. After obtaining the $U_{data}$ of each sample, we split $\mathcal{S}$ into $M$ datasets $\mathcal{D}=\{\mathcal{D}^{m}\}_{m=1}^{M}$ according to $U_{data}$ and sort $\mathcal{D}$ from small $U_{data}$ to large $U_{data}$. So far, we get the data learning order for the model. The model is trained to converge on each distribution $\mathcal{D}^{m}$, which can be regarded as the model solving $M$ tasks $\mathcal{M}=\{\mathcal{M}^{m}\}_{m=1}^{M}$.

\begin{itemize}[leftmargin=10 pt]
    \item The \textit{objective-confidence scheduling module} controls the overall learning process of each task $\mathcal{M}$ and dataset $\mathcal{D}$. It determines the new data to learn and the old data to review based on the objective-confidence and aims to solve problems of catastrophic forgetting and overfitting. It imitates the fact that students will decide what to review based on their test scores.
\end{itemize}

This module is mainly controlled by the confidence measurer as shown in the yellow part of Figure \ref{fig:method}. The confidence measurer organizes and arranges datasets $\mathcal{D}=\{\mathcal{D}^{m}\}_{m=1}^{M}$, which is initially obtained from data arrangement module. After the model learning the task $\mathcal{M}^{m}$ with dataset $\mathcal{D}^{m}$, the confidence measurer will determine the dataset $\mathcal{D}^{m+1}$ that the model will learn next. $\mathcal{D}^{m+1}$ contains samples in new distribution and samples that should be re-learned. The confidence measurer finds samples to review according to objective-confidence in Equation \ref{eq:importance}. Sample with larger $\beta$ need to be learned again. The partial relay will alleviate the catastrophic forgetting without causing the over fitting caused by over training.

\begin{itemize}[leftmargin=10 pt]
    \item The \textit{self-confidence scheduling module} controls the duration of each training stage with a task $\mathcal{M}^{m}$. It determines the training direction of the model on the dataset $\mathcal{D}^{m}$ and the evidence of model convergence by the self-confidence method. It imitates the fact that students decide whether they have mastered the knowledge or not through their confidence in the current knowledge.
\end{itemize}

This module is mainly controlled by the confidence scheduler as shown in the gray part of Figure \ref{fig:method}. The confidence scheduler determines whether the model converges on the current dataset $\mathcal{D}^{m}$. If the model converges, it will obtain $\mathcal{D}^{m+1}$ provided by the confidence measurer and let the model start learning the new task $\mathcal{M}^{m+1}$; If not, it will continue training the model on the current dataset $\mathcal{D}^{m}$. The confidence scheduler judges whether the model converges according to Equation \ref{eq:model confidence}. Training stops when confidence is high. Specifically, like the early stop based on the loss \cite{DBLP:conf/nips/CaruanaLG00}, the model is stopped training if the self-confidence is no longer increased in several epochs.

The overall algorithm is in Algorithm \ref{alg:C3TS}. It first pre-train the model and get the initial $\mathcal{M}$ and $\mathcal{D}$. Then, it trains the model from $\mathcal{M}^{1}$ to $\mathcal{M}^{M}$. For each $\mathcal{M}^{m}$, it controls the training duration based on the self-confidence. Between $\mathcal{M}^{m}$ and $\mathcal{M}^{m+1}$, it dynamically adjusts $\mathcal{D}^{m+1}$ according to the objective-confidence.

\subsection{Complexity Analysis}

Assuming that the computational complexity of updating a neural network model by one sample is $\mathcal{O}(d)$, then training a model by $D$ time series with length $T$ and $E$ epochs usually costs $\mathcal{O}(TEDd)$. \mname contains a pre-training process and a training process. The pre-training process with loss in Equation \ref{eq:importance} has the complexity of $\mathcal{O}(TEDd)$, being the same as the normal method. In the training process, \mname trains a model with $N$ distributions $\mathcal{D}$ with $N$ training tasks $\mathcal{M}$, assuming there are $E'$ epochs and $S$ retained sample in each $\mathcal{M}^{n}$, the complexity will be $\mathcal{O}(NE'(D+S)d)$. The overall complexity is $\mathcal{O}((TE+NE')(2D+S)d)$. As $S<<D$ and $N$ is a small constant with $N<T$, the complexity of \mname approximates $\mathcal{O}(TEDd)$, almost being the same as the general training strategy.

\begin{table*}[!t]
\caption{Baselines Classification Accuracy (AUC-ROC$\uparrow$) for 4 Real-world Datasets at 10 Time Steps.\newline
\footnotesize {*k\% means the current classification is based on k\% of the length of the time series;
Bold font indicates the highest accuracy.}}\label{tb:accuracy}
\centerline{
\small
\setlength{\tabcolsep}{0.33mm}{
\begin{tabular}{llcccccccccc}
\toprule[0.8pt]
 Dataset &Method    &10\%*  &20\%   &30\%    &40\%  &50\%   &60\% &70\%  &80\%    &90\%   &100\%  \\
\midrule[0.8pt]
\multirow{8}*{UCR-EQ} 
& SR  &0.711±0.015 &0.736±0.014     &0.802±0.018     &0.863±0.015   &0.879±0.012    &0.888±0.017 &0.920±0.015     &0.928±0.005   &0.938±0.008    &0.941±0.004       \\
& ECEC &0.710±0.013 &0.738±0.018    &0.809±0.015    &0.865±0.014  &0.881±0.013  &0.890±0.015       &0.922±0.014  &0.929±0.007   &0.937±0.010  &0.940±0.009       \\
& CLEAR  &0.725±0.011 &0.768±0.018   &0.821±0.017    &0.874±0.016   &0.882±0.009    &0.895±0.014     &0.919±0.008  &0.923±0.002 &0.930±0.010 &0.933±0.003       \\
& CLOPS  &0.728±0.013  &0.767±0.017  &0.819±0.013  &0.876±0.016 &0.877±0.014  &0.885±0.012   &0.920±0.015       &0.929±0.008    &0.932±0.007  &0.934±0.004       \\
& DIF  &0.724±0.018  &0.770±0.015   &0.825±0.015  &0.880±0.013   &0.886±0.013   &0.904±0.012   &0.913±0.014  &0.923±0.004  &0.929±0.008  &0.932±0.005       \\
& UNCERT  &0.720±0.012 &0.773±0.016   &0.821±0.016  &0.878±0.016  &0.886±0.012 &0.902±0.015       &0.908±0.011 &0.917±0.006   &0.921±0.011  &0.925±0.005       \\
& DROPOUT  &0.721±0.014  &0.765±0.017  &0.823±0.013  &0.870±0.019   &0.883±0.010   &0.901±0.011    &0.905±0.013   &0.920±0.010    &0.923±0.009  &0.930±0.011       \\
& TCP  &0.729±0.011  &0.771±0.012   &0.827±0.012  &0.881±0.010  &0.881±0.011   &0.898±0.015     &0.907±0.010 &0.921±0.009 &0.925±0.010   &0.933±0.008       \\
& STL  &0.731±0.017 &0.770±0.013   &0.830±0.017  &0.875±0.012  &0.879±0.008   &0.906±0.005     &0.910±0.009 &0.918±0.006   &0.928±0.007  &0.934±0.005       \\
&\textbf{\mname}  &\textbf{0.736±0.013} &\textbf{0.774±0.014}  &\textbf{0.840±0.014}   &\textbf{0.882±0.014}    &\textbf{0.899±0.010}      &\textbf{0.907±0.009} &\textbf{0.925±0.010}   &\textbf{0.933±0.008}   &\textbf{0.939±0.006}  &\textbf{0.946±0.003}   \\

\midrule[0.4pt]
\multirow{8}*{USHCN}
& SR    &0.719±0.020  &0.730±0.022   &0.750±0.020     &0.761±0.023    &0.802±0.020  &0.836±0.016     &0.889±0.012    &0.902±0.013    &0.923±0.010     &0.933±0.009  \\
& ECEC  &0.721±0.019  &0.736±0.024       &0.752±0.021  &0.760±0.025   &0.800±0.019    &0.837±0.016    &0.890±0.011    &0.906±0.017  &0.921±0.011    &0.931±0.009  \\
& CLEAR  &0.720±0.022  &0.736±0.025     &0.770±0.023  &0.798±0.024  &0.801±0.020 &0.834±0.016     &0.883±0.016 &0.896±0.017    &0.914±0.011     &0.926±0.007  \\
& CLOPS  &0.722±0.025  &0.728±0.026      &0.758±0.024  &0.781±0.023    &0.805±0.018  &0.838±0.013   &0.885±0.013   &0.899±0.010    &0.923±0.009   &0.928±0.005  \\
& DIF   &0.725±0.021  &0.738±0.025     &0.756±0.023   &0.784±0.024    &0.798±0.020 &0.837±0.010    &0.875±0.011  &0.879±0.012 &0.912±0.010    &0.921±0.004  \\
& UNCERT  &0.724±0.023  &0.740±0.024      &0.751±0.019    &0.781±0.025   &0.800±0.023   &0.835±0.016   &0.873±0.012   &0.877±0.011  &0.907±0.007    &0.919±0.007  \\
& DROPOUT &0.719±0.017  &0.738±0.016     &0.750±0.015   &0.782±0.014   &0.797±0.018   &0.829±0.013   &0.877±0.013   &0.885±0.012   &0.908±0.013   &0.920±0.015  \\
& TCP   &0.722±0.016  &0.738±0.012      &0.756±0.018  &0.785±0.013   &0.806±0.017  &0.839±0.011     &0.895±0.011 &0.900±0.011  &0.920±0.011   &0.929±0.014  \\
& STL  &0.718±0.015   &0.730±0.016       &0.757±0.017   &0.788±0.015    &0.802±0.016  &0.836±0.017   &0.871±0.011   &0.876±0.013   &0.916±0.012   &0.925±0.013  \\
&\textbf{\mname} &\textbf{0.728±0.016} &\textbf{0.742±0.017}  &\textbf{0.759±0.018}     &\textbf{0.791±0.019}   &\textbf{0.811±0.017}   &\textbf{0.841±0.012}   &\textbf{0.897±0.014 }    &\textbf{0.910±0.014 }   &\textbf{0.930±0.012}   &\textbf{0.937±0.013}\\

\midrule[0.4pt]
\multirow{8}*{COVID-19\ } 
& SR  &0.730±0.024   &0.831±0.022     &0.867±0.016    &0.900±0.018 &0.913±0.013   &0.923±0.010  &0.937±0.007    &0.946±0.006  &0.960±0.004   &0.962±0.005   \\
& ECEC   &0.732±0.028   &0.834±0.020     &0.870±0.016  &0.904±0.014 &0.913±0.012  &0.924±0.015   &0.940±0.012 &0.948±0.007  &0.958±0.008  &0.963±0.007   \\
& CLEAR    &0.769±0.015   &0.837±0.019 &0.875±0.019  &0.888±0.018   &0.916±0.017 &0.923±0.014     &0.936±0.012 &0.940±0.013   &0.955±0.009  &0.956±0.008      \\
& CLOPS    &0.779±0.017  &0.835±0.018  &0.869±0.015  &0.885±0.021   &0.914±0.017   &0.924±0.017    &0.936±0.011  &0.939±0.010  &0.951±0.007  &0.953±0.005      \\
& DIF    &0.785±0.019  &0.830±0.021  &0.862±0.017  &0.879±0.016  &0.915±0.014  &0.926±0.014       &0.936±0.010 &0.941±0.007 &0.947±0.006   &0.952±0.006       \\
& UNCERT   &0.775±0.013  &0.841±0.016  &0.871±0.015  &0.900±0.013  &0.915±0.013 &0.925±0.015     &0.935±0.009 &0.940±0.007  &0.950±0.005   &0.954±0.006       \\
& DROPOUT    &0.740±0.017  &0.831±0.020 &0.860±0.013   &0.885±0.012  &0.912±0.011    &0.924±0.018   &0.932±0.011   &0.939±0.007  &0.943±0.004  &0.945±0.005      \\
& TCP    &0.786±0.012   &0.832±0.019  &0.872±0.018  &0.895±0.016   &0.915±0.014 &0.927±0.011       &0.941±0.010 &0.942±0.007   &0.948±0.004 &0.947±0.008       \\
& STL   &0.770±0.012   &0.833±0.017  &0.870±0.011 &0.895±0.014  &0.916±0.013 &0.925±0.013     &0.941±0.012 &0.944±0.011   &0.948±0.007 &0.950±0.008       \\
&\textbf{\mname}  &\textbf{0.790±0.021}  &\textbf{0.843±0.19}   &\textbf{0.877±0.020}  &\textbf{0.901±0.015}  &\textbf{0.919±0.015}   &\textbf{0.927±0.012}   &\textbf{0.945±0.011}   &\textbf{0.960±0.011}   &\textbf{0.967±0.010}  &\textbf{0.969±0.010} \\

\midrule[0.4pt]
\multirow{8}*{SEPSIS} 
& SR  &0.652±0.017  &0.701±0.016   &0.728±0.017  &0.749±0.025   &0.821±0.026 &0.825±0.027    &0.827±0.020  &0.837±0.018 &0.845±0.014      &0.866±0.023     \\
& ECEC   &0.655±0.013  &0.702±0.014  &0.731±0.019  &0.752±0.025   &0.823±0.016   &0.825±0.019  &0.829±0.014     &0.839±0.015 &0.849±0.016      &0.863±0.014     \\
& CLEAR  &0.667±0.022 &0.711±0.022  &0.733±0.023    &0.763±0.020    &0.827±0.026    &0.830±0.026    &0.838±0.024  &0.847±0.017     &0.848±0.015     &0.854±0.016    \\
& CLOPS   &0.665±0.026 &0.709±0.023  &0.730±0.024   &0.765±0.025     &0.826±0.023  &0.831±0.025    &0.836±0.028   &0.846±0.015   &0.849±0.014   &0.853±0.012    \\
& DIF  &0.666±0.025 &0.710±0.021  &0.732±0.024   &0.755±0.027    &0.825±0.025   &0.832±0.027   &0.839±0.028      &0.844±0.012 &0.847±0.010      &0.848±0.016    \\
& UNCERT  &0.665±0.026 &0.705±0.019  &0.733±0.025  &0.759±0.028     &0.824±0.029   &0.831±0.027   &0.838±0.026   &0.846±0.015    &0.850±0.017    &0.857±0.018    \\
& DROPOUT  &0.660±0.021 &0.766±0.015  &0.720±0.014    &0.748±0.021    &0.820±0.020     &0.825±0.022  &0.832±0.021  &0.835±0.018    &0.840±0.015   &0.850±0.011    \\
& TCP  &0.662±0.021 &0.705±0.016  &0.722±0.020   &0.758±0.025    &0.826±0.027    &0.827±0.027  &0.839±0.025      &0.840±0.017 &0.843±0.011      &0.862±0.016    \\
& STL   &0.660±0.023 &0.709±0.016  &0.727±0.021    &0.760±0.024   &0.824±0.026     &0.833±0.028      &0.836±0.015 &0.840±0.016  &0.845±0.016    &0.855±0.017    \\
&\textbf{\mname} &\textbf{0.671±0.025}   &\textbf{0.715±0.024} &\textbf{0.734±0.021}   &\textbf{0.768±0.026}   &\textbf{0.831±0.023}       &\textbf{0.840±0.024}    &\textbf{0.840±0.018}  &\textbf{0.851±0.012}  &\textbf{0.853±0.012}    & \textbf{0.872±0.012}  \\
\bottomrule[0.8pt]
\end{tabular}
}}
\end{table*}

\section{Experiments}
More detailed experiments and analyses are in Appendix.

\subsection{Datasets}
\begin{itemize} [leftmargin=10 pt]
    \item UCR-EQ dataset \cite{UCRArchive} has 471 earthquake records from UCR time series database archive. It is the univariate time series of seismic feature value. Natural disaster early warning, like earthquake warning, helps to reduce casualties and property losses \cite{2021Earthquake}.
    
    \item USHCN dataset \cite{USHCN} has the daily meteorological data of 48 states in U.S. from 1887 to 2014. It is the multivariate time series of 5 weather features. Rainfall warning is not only the demand of daily life, but also can help prevent natural disasters \cite{2021The}. 
     
    \item COVID-19 dataset \cite{COVID-19} has 6,877 blood samples of 485 COVID-19 patients from Tongji Hospital, Wuhan, China. It is the multivariate time series of 74 laboratory test features. Mortality prediction helps for the personalized treatment and resource allocation \cite{DBLP:journals/BMC/sun}. 
    
    \item SEPSIS dataset \cite{DBLP:conf/cinc/ReynaJSJSWSNC19} has 30,336 patients' records, including 2,359 diagnosed sepsis. It is the multivariate time series of 40 related patient features. Early diagnose of sepsis is critical to improve the outcome of ICU patients \cite{seymour2017time}.
\end{itemize}
Not that for each time series in the above four datasets, every time point is tagged with a class label, which is the same as its outcome, such as earthquake, rain, mortality, sepsis.

\subsection{Baselines}

Based on Section \ref{sec:related work}, we use four types of baselines. They are all training strategies, so we use the same LSTM model for them.

\noindent The first is early classification of time series (ECTS-based).

\begin{itemize}[leftmargin=10 pt]
\item SR \cite{DBLP:journals/tnn/MoriMDL18}. It has multiple models trained by the full-length time series. The classification result is the fusion result of its models.

\item ECEC \cite{lv2019effective}. It has multiple models. Different models are trained by the data in different time stages. When classifying, it selects the classifier based on the time stage of the input data.
\end{itemize}

\noindent The second is replay continual learning (Replay-based).

\begin{itemize} [leftmargin=10 pt]
\item CLEAR \cite{DBLP:conf/nips/RolnickASLW19}. It uses the reservoir sampling to limit the number of stored samples to a fixed budget assuming an i.i.d. data stream.

\item CLOPS \cite{2021A}. It trains a base model by replaying old tasks with importance-guided buffer storage to avoid catastrophic forgetting.
\end{itemize}

\noindent The third is curriculum learning (CL-based).

\begin{itemize} [leftmargin=10 pt]
\item DIF \cite{DBLP:conf/icml/HacohenW19}. It arranges curriculum (data learning order/task order) by loss/difficulty getting from the teacher model.

\item UNCERT \cite{2021A}. It arranges curriculum (data learning order/task order) by sentence uncertainty and model uncertainty.
\end{itemize}

\noindent The fourth is about the confidence research (Confidence-based).

\begin{itemize} [leftmargin=10 pt]
\item DROPOUT \cite{DBLP:conf/icml/GalG16}. It uses the Monte Carlo Dropout method to approximate Bayesian inference and gets the model uncertainty.

\item TCP \cite{DBLP:conf/nips/CorbiereTBCP19}. It designs a true class probability to estimate the confidence of the model prediction.

\item STL \cite{DBLP:journals/corr/abs-2105-11545}. It gives the uncertainty about the signal temporal logic, which is more suitable for time series.
\end{itemize}

\begin{table*}[!h]
\caption{The Performances of Solving Catastrophic Forgetting Problem. \\
\footnotesize{The left part is BWT$\uparrow$ results of baselines, the right part is FWT$\uparrow$ results of baselines.}}\label{tb:BWTFWT}
\centerline{
\small
\setlength{\tabcolsep}{1.6mm}{
\begin{tabular}{l|ccccccc|ccccccc}
\toprule[0.8pt]
\tiny{\diagbox{Dataset}{Method}}   &SR  &ECEC  &CLEAR    &CLOPS  &DIF   &STL  &\textbf{\mname} &SR  &ECEC  &CLEAR    &CLOPS &DIF   &STL    &\textbf{\mname} \\
\midrule[0.8pt]
UCR-EQ      &+0.019 &+0.021   &+0.053 &+0.052 &+0.022 &+0.020   &\textbf{+0.058}   &+0.121  &+0.129   &+0.312 &+0.301 &+0.124  &+0.120  &\textbf{+0.345}\\
USHCN       &+0.028 &+0.034   &+0.063 &+0.074 &+0.017 &+0.023 &\textbf{+0.084}   &+0.212   &+0.128   &+0.335  &+0.301  &+0.205 &+0.216 &\textbf{+0.342} \\
COVID-19     &-0.001 &+0.010   &+0.009  &+0.014  &+0.002 &+0.006   &\textbf{+0.020}   &+0.126    &+0.221 &+0.427  &+0.439 &+0.220    &+0.232  &\textbf{+0.455}  \\
SEPSIS    &-0.019 &-0.017   &+0.030  &+0.032  &-0.010 &-0.008   &\textbf{+0.035}   &+0.095   &+0.165 &+0.401  &+0.397    &+0.106   &+0.124 &\textbf{+0.410}      \\
\bottomrule[0.8pt]
\end{tabular}
}}
\end{table*}

\begin{table*}[!h]
\caption{The Performances of Solving Over Fitting Problem.\\
\footnotesize {Sepsis classification accuracy with non-uniform training sets and validation sets. $\downarrow$ means the accuracy is greatly reduced (top 3).}} \label{tb:overfitting}
\centerline{
\small
\setlength{\tabcolsep}{0.82mm}{
\begin{tabular}{l|ll|ll|ll|lll|l}
\toprule[0.8pt]
Subset   &SR  &ECEC    &CLEAR    &CLOPS    &DIF  &UNCERT  &\textbf{DROPOUT} &\textbf{TCP} &\textbf{STL}   &\textbf{\mname}   \\
\midrule[0.8pt]
Male   &0.844±0.017   &0.851±0.019   &0.853±0.017 &0.855±0.016 &0.852±0.015  &0.858±0.024   &0.853±0.022 &0.850±0.021 &0.856±0.022  &0.860±0.022\\
Female    &0.817±0.026$\downarrow$   &0.827±0.024   &0.835±0.024 &0.830±0.025$\downarrow$  &0.832±0.021  &0.823±0.021$\downarrow$   &0.849±0.019 &0.848±0.021 &0.850±0.020 &0.853±0.021\\
\midrule[0.4pt]
30-   &0.845±0.021   &0.853±0.016   &0.863±0.012 &0.867±0.016 &0.861±0.015  &0.854±0.024 &0.848±0.021 &0.850±0.022 &0.852±0.024   &0.864±0.022\\
30+    &0.814±0.026  &0.812±0.024$\downarrow$    &0.820±0.025$\downarrow$ &0.824±0.026  &0.809±0.027$\downarrow$  &0.838±0.024   &0.838±0.021 &0.840±0.020 &0.832±0.023  &0.843±0.025\\
\bottomrule[0.8pt]
\end{tabular}
}}
\end{table*}

\begin{table*}[!h]
\caption{Model Performances (Result Accuracy, Model Convergence, Result Uncertainty) under Different Learning Order.\\
\footnotesize {All strategies train a same LSTM model to classify Sepsis, and their difference is reflected in the data learning order. Accuracy $\uparrow$ is evaluated by using the full-length time series; Epochs $\downarrow$ is the training rounds when the model converges; $\alpha \downarrow$ is the expected non-coverage probability for results.}} \label{tb:training order}
\centerline{
\small
\setlength{\tabcolsep}{1.8mm}{
\begin{tabular}{lrrrrrrrrr}
\toprule[0.8pt]
  &Random &Time$\uparrow$  &Time$\downarrow$    &Difficulty$\uparrow$   &Difficulty$\downarrow$    &Uncertainty$\uparrow$  &Uncertainty$\downarrow$   &Confidence$\uparrow$ &\textbf{Confidence$\downarrow$}   \\
\midrule[0.8pt]
Accuracy  &0.831±0.018 &0.854±0.013   &0.861±0.012   &0.862±0.015 &0.852±0.014 &0.866±0.017  &0.853±0.022   &0.853±0.017 &\textbf{0.872±0.016} \\
Epochs  &637  &581 &532 &496 &525 &475 &558  &546 &\textbf{457}\\
$\alpha$ &0.88 &0.67 &0.71 &0.65 &0.67 &\textbf{0.45} &0.52  &0.54 &0.51\\
\bottomrule[0.8pt]
\end{tabular}
}}
\end{table*}

\subsection{Evaluation Metrics}

The classification accuracy is evaluated by the Area Under Curve of Receiver Operating Characteristic (AUC-ROC). The performance of solving forgetting\&overfitting is evaluated by Backward Transfer (BWT) and Forward Transfer (FWT), the influence that learning a current has on the old/future. $R\in \mathbb{R}^{|\mathcal{M}|\times |\mathcal{M}|}$ is an accuracy matrix, $R_{i,j}$ is the accuracy on $\mathcal{M}^{j}$ after learning $\mathcal{M}^{i}$. $\overline{b}$ is the accuracy with random initialization.
\begin{equation} 
BWT=\frac{1}{|\mathcal{M}|-1}\sum_{i=1}^{|\mathcal{M}|-1}R_{|\mathcal{M}|,i}-R_{i,i}
\end{equation}
\begin{equation} 
FWT=\frac{1}{|\mathcal{M}|-1}\sum_{i=2}^{|\mathcal{M}|}R_{i-1,i}-\overline{b}_{i,i}
\end{equation}

The result uncertainty is evaluated by prediction intervals (PI). We use the conditional PI specifically \cite{DBLP:journals/asc/KabirKKNS21}. $CP$ is the conditional probability function that can construct an interval of $(1-\alpha)$ confidence level. $\alpha$ is the expected non-coverage probability.
\begin{equation} 
PI=[CP^{-1}(\frac{\alpha}{2}),CP^{-1}(1-\frac{\alpha}{2})]
\end{equation}

\subsection{Results and Analysis}
\subsubsection{Analysis of multi-distribution}\quad

Before discussing method performances, we show the basic scenario of CCTS: multi-distribution. The data in different time stages have distinct statistical characteristics and finally form multiple distributions. Comparing Figure \ref{fig:introduction}, Figure \ref{fig:multi-distribution} and Figure \ref{fig:sepsis distribution}, different data arrangement will lead to different multi-distribution. The fundamental goal of the following experiment is to model them.

\begin{table}[!h]
\small
\centering
\caption{The percentage (\%) of different samples between \mname and other methods in each data distribution of 4 training steps }\label{tb:difference}
\setlength{\tabcolsep}{8mm}{
\begin{tabular}{lrrrrr}
\toprule[0.8pt]
 Training / baby step  &1 &\textbf{2} &\textbf{3}  &\textbf{4}   &5 \\
\midrule[0.8pt]
Time$\downarrow$ &0.14 &0.24 &0.27 &0.20 &0.10\\
Difficulty$\uparrow$ (DIF) &0.17 &0.28 &0.22 &0.18 &0.09\\
Uncertainty$\uparrow$ (UNCERT) &0.13 &0.25 &0.23 &0.16 &0.08\\
\bottomrule[0.8pt]
\end{tabular}}
\end{table}

\begin{table}[!h]
\small
\centering
\caption{\mname Performance with Different Distribution Number}\label{tb:datasetnumber}
\setlength{\tabcolsep}{6mm}{
\begin{tabular}{lrrrrr}
\toprule[0.8pt]
 N  &2    &4    &6   &8    &10    \\
\midrule[0.8pt]
Accuracy    &0.839±0.01    &\textbf{0.861±0.01}  &0.816±0.01   &0.797±0.01 &0.784±0.01  \\
BWT  &+0.153    &\textbf{+0.156}  &+0.139    &+0.135 &+0.125    \\
FWT &+0.398    &\textbf{+0.424}  &+0.379   &+0.365 &+0.364  \\
Batches &685    &657  &632   &\textbf{578}   &594 \\
\bottomrule[0.8pt]
\end{tabular}}
\end{table}

\begin{table}[!h]
\small
\centering
\caption{\mname Performance with Different Retrained Samples}\label{tb:replaynumber}
\setlength{\tabcolsep}{6mm}{
\begin{tabular}{lrrrrr}
\toprule[0.8pt]
 M  &1    &2    &3   &4    &5    \\
\midrule[0.8pt]
Accuracy    &\textbf{0.861±0.01}    &0.841±0.01  &0.826±0.01   &0.813±0.01 &0.815±0.01  \\
BWT  &\textbf{+0.156}    &+0.144  &+0.127    &+0.130 &+0.128    \\
FWT &+0.424    &\textbf{0.443}  &+0.389   &+0.376 &+0.368  \\
\bottomrule[0.8pt]
\end{tabular}}
\end{table}

\begin{figure}[!h]
\caption{Catastrophic Forgetting and Over Fitting Case.\\
\footnotesize{Accuracy decreases on Sepsis distribution 1 after learning Sepsis distribution 2. Much retraining on distribution 1 lets model over fit in it. While retraining partial samples of distribution 1 can alleviate these two problems.}} \label{fig:introduction3}
\centering
\includegraphics[width=0.6\linewidth]{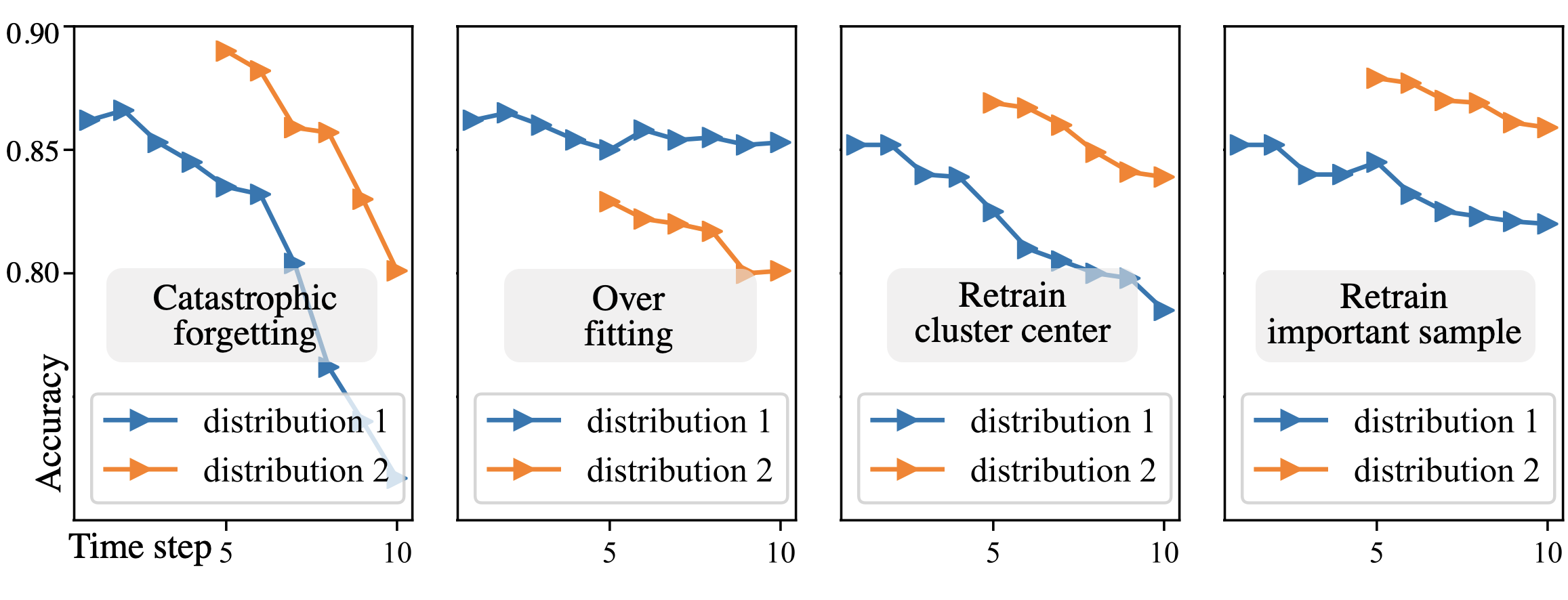}
\end{figure} 

\begin{figure}[!h]
\caption{Model Performances under Different Training Orders.\\
\footnotesize{For a LSTM model, its Sepsis classification accuracy and average training epochs are different by training orders of early-to-later, later-to-early, confidence-increasing and confidence-decreasing.} }\label{fig:introduction4}
\centering
\includegraphics[width=0.6\linewidth]{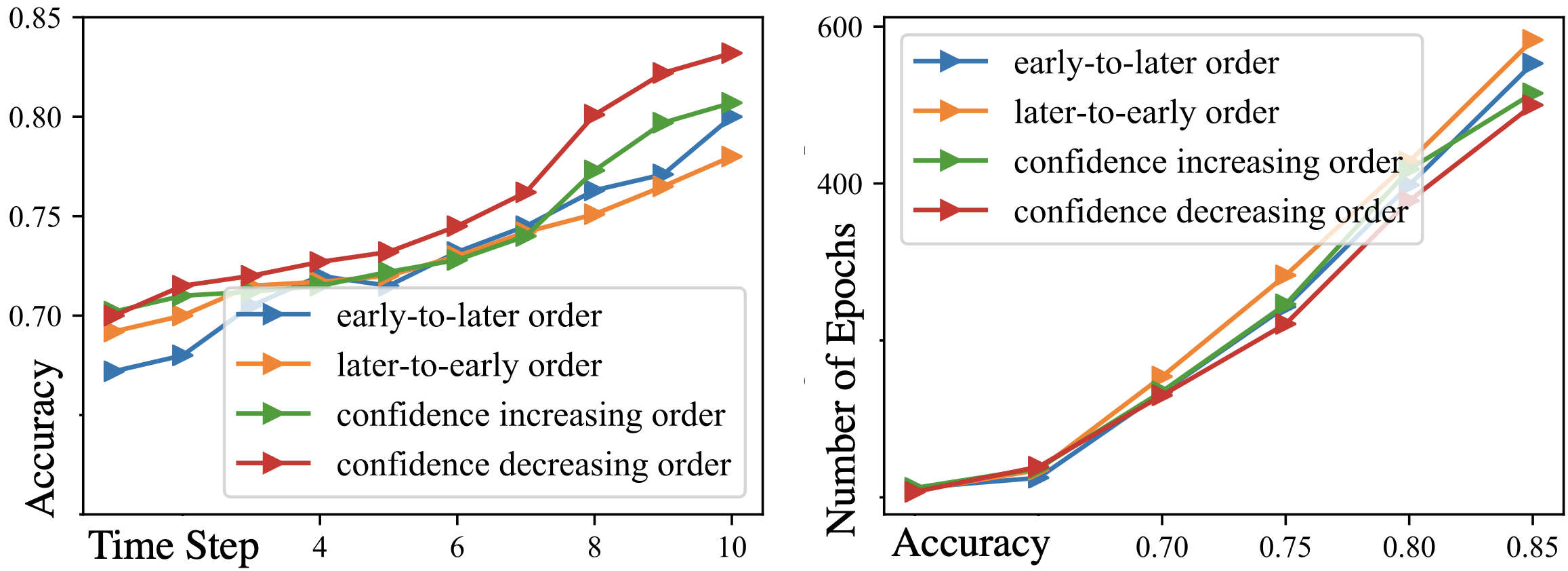}
\end{figure}

\subsubsection{Continuous classification accuracy}\quad

Our method \mname is significantly better than all baselines. In Bonferroni-Dunn test, as shown in Table \ref{tb:accuracy}, $k=10$, $n=4$, $m=5$ are the number of methods, datasets, cross-validation fold, then $N=n\times m=20$, correspondingly,$ q_{0.05}=2.773$ and $ \textit{CD}=q_{0.05}\sqrt{\frac{k(k+1)}{6N}}=2.655$, finally $\textit{rank(CCTS)}=1<\textit{CD}$. Thus, the accuracy is significantly improved. Specifically, CCTS can classify more accurately at every time than 9 baselines. Take sepsis diagnosis as an example, compared with the best baseline, our method improves the accuracy by 1.2\% on average, 2.1\% in the early 50\% time stage when the key features are unobvious. Each hour of delayed treatment increases sepsis mortality by 4-8\% \cite{seymour2017time}. With the same accuracy, we can predict 0.965 hour in advance.

\subsubsection{Difficulty 1: Catastrophic forgetting and over fitting}\quad

Our method \mname can alleviate this problem. As shown in Table \ref{tb:accuracy}, \mname has the best performance on the early time series, showing the ability to alleviate catastrophic forgetting. As shown in Table \ref{tb:BWTFWT}, \mname has the highest BWT, meaning it has the lowest negative influence that learning the new tasks has on the old tasks. \mname has the highest FWT, meaning it has the highest positive influence that learning the former data distributions has on the task, especially for Sepsis and COVID-19 datasets. Figure \ref{fig:introduction3} shows the case study that the partial replay of importance samples trades off forgetting and overfitting. Meanwhile, our method can avoid model overfitting and guarantee certain model generalizations. In Table \ref{tb:overfitting}, for most baselines, the accuracy on the validation set is much lower than that on the training set. But confidence-based methods, DROPOUT, TCP, STL and \mname, have relatively better generalization performance, which shows the potential of confidence-guided training strategies.

\subsubsection{Difficulty 2: Optimal data learning order}\quad

The learning order based on confidence decline makes the LSTM model perform best in classification accuracy and model convergence, and perform well in result uncertainty, shown in Table \ref{tb:training order} and Figure \ref{fig:introduction4}. Most existing methods ignore the data learning order and train the model randomly. Nowadays, some work \cite{DBLP:journals/corr/abs-2010-13166} has paid attention to the data learning order, but the existing approaches basically use the difficulty and uncertainty to measure data. Experiments show that it has greater potential to define the order of model learning data by imitating human confidence in knowledge. Internal differences among these methods are analyzed in the next section.

\begin{figure}[t]
\caption{Multi-distribution in Sepsis Dataset}
\centering
\includegraphics[width=0.5\linewidth]{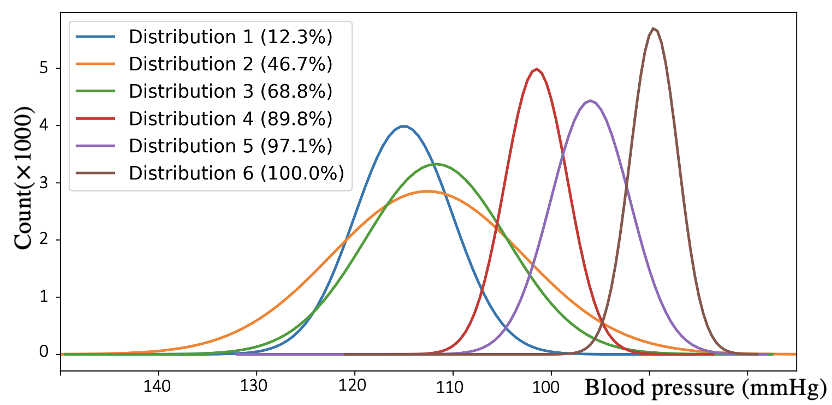}
\label{fig:multi-distribution}
\end{figure} 

\begin{figure}[t]
\caption{Four Distributions in Sepsis Dataset}
\centering
\includegraphics[width=0.5\linewidth]{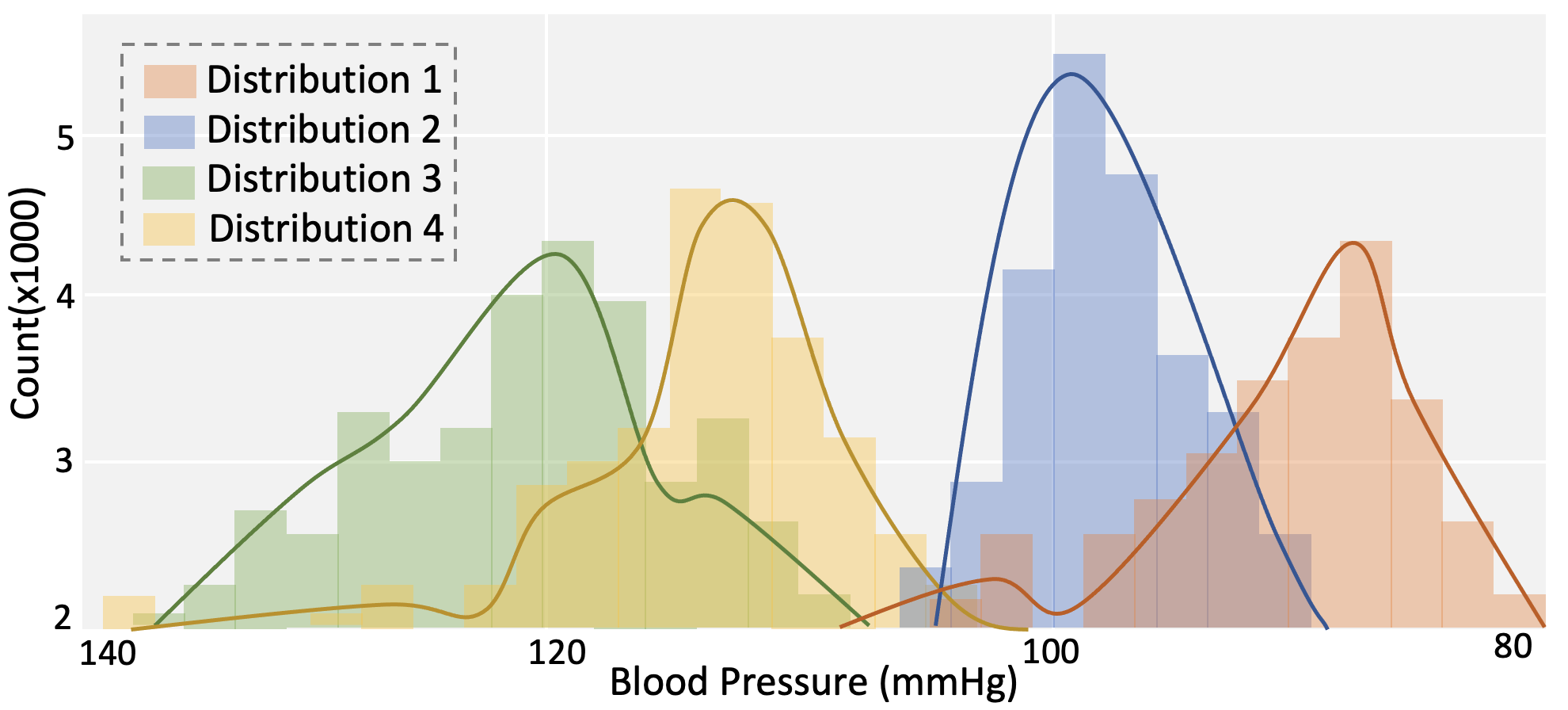}
\label{fig:sepsis distribution}
\end{figure} 

\subsubsection{The difference among baselines}\quad

\mname and baselines will divide data into different distributions training/baby steps. Their divisions are considerably different. As shown in Table \ref{tb:difference}, the difference in the middle step, steps 2-4, is greater than that in steps 1 and 5. It shows that the most simple and complex time series quantified by different measures are relatively similar, and the main diversity lies in those samples of which the difficulties hardly to be distinguished. Therefore, we argue that the improvements of \mname may be mainly contributed by the differences in these middle steps.

\subsubsection{Analysis of hyper-parameters}\quad

Different numbers of data distributions or baby steps will result in different model performances. We first construct $U_{data}$ value of training data into a normal distribution $\mathcal{N}_{u}(\mu,\sigma^{2})$, then use its confidence intervals $-n\sigma<\mu<-(n-1)\sigma$ to split the dataset and get initial $N$ baby steps. As shown in Table \ref{tb:datasetnumber}, for sepsis classification, the best number of distribution or baby step is 4. We show these 4 distributions by visualizing blood pressure in Figure \ref{fig:sepsis distribution}, which is obviously different from 6 distributions divided 
by time stages in Figure \ref{fig:multi-distribution}. Meanwhile, Different replayed samples will also result in different model performances. We construct the importance coefficient $\beta$ into a positive skewed distribution $\mathcal{N}_{\beta}(m_{o},\sigma^{2})$ with $m_{o}<m_{e}<m$ and make $\epsilon=m$. The important sample is the data with $\beta>\epsilon$. To test the influence of the number of important samples, we replay $\frac{1}{M}$ of the data with larger $\beta$. For Sepsis dataset, the result is better with the decrease of $M$.

\begin{figure}[t]
\caption{Replayed Important Samples}
\centering
\includegraphics[width=0.6\linewidth]{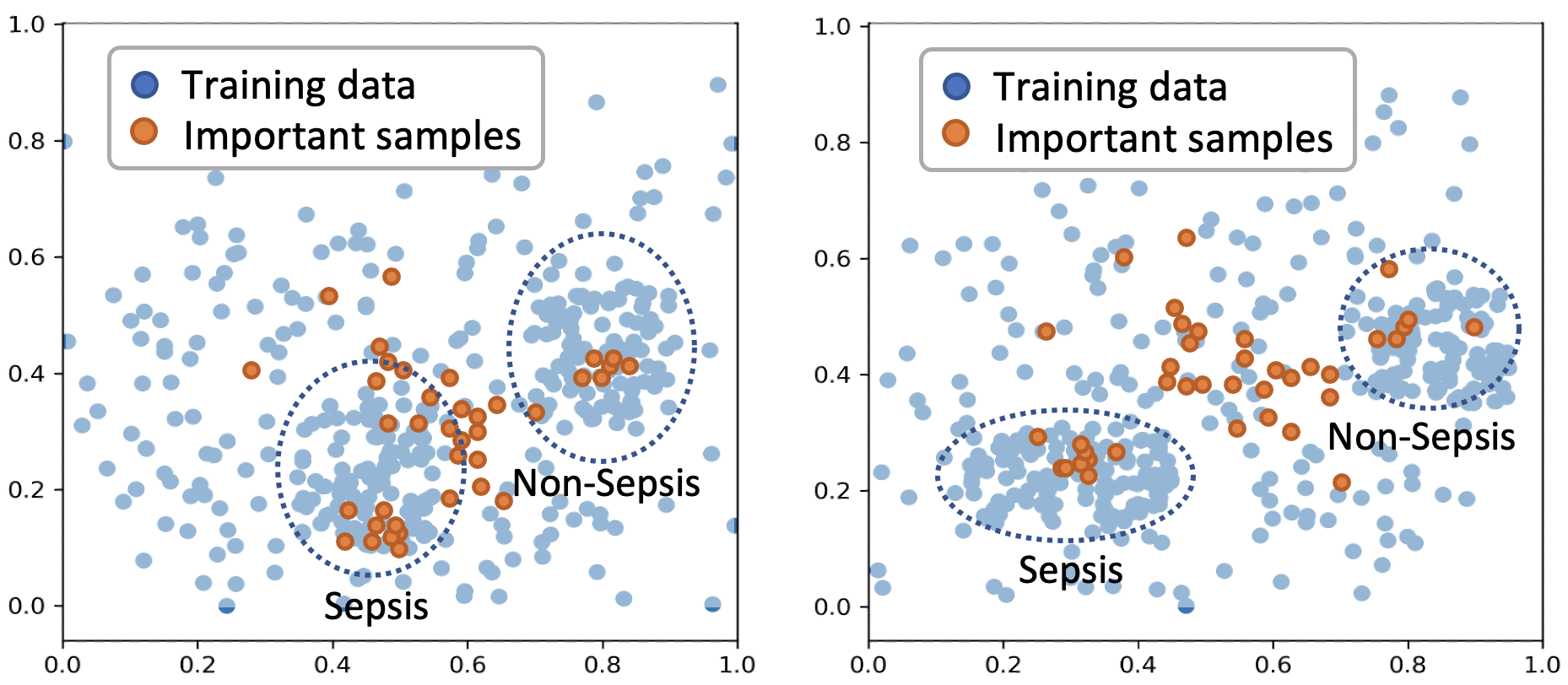}
\label{fig:replay sample}
\end{figure} 

\begin{figure}[t]
\centering
\caption{Curves of Confidence at Different Checkpoints}
\includegraphics[width=0.5\linewidth]{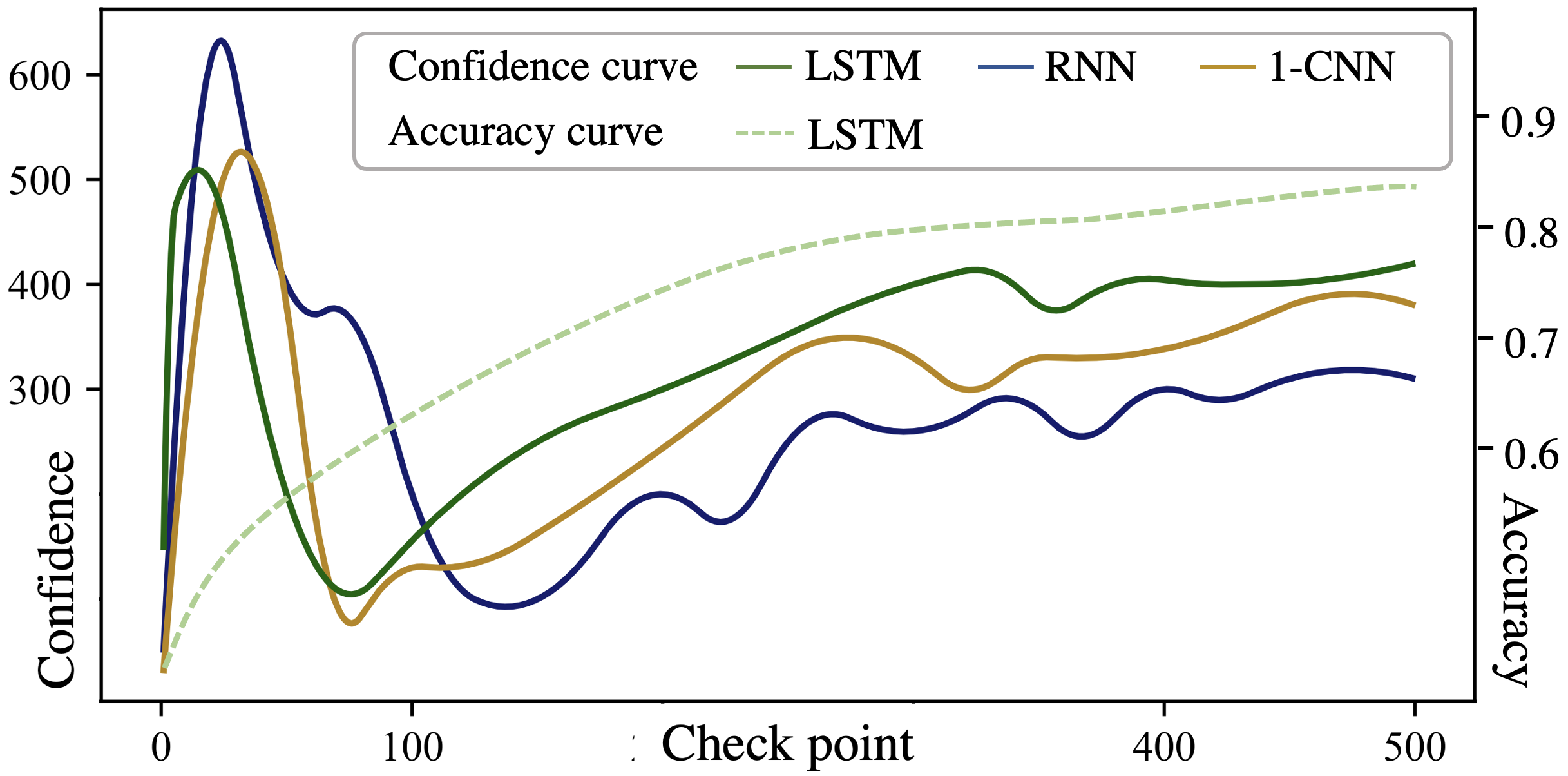}
\label{fig:confidence change}
\end{figure}

\subsubsection{Analysis of Replayed Important Data}\quad

The important samples include not only the data hard to learn but also the representative data in each class, shown in Figure \ref{fig:replay sample}. It might be because the representative data is similar to the most common data, resulting in a greater additive loss, therefore leading to smaller coefficients in Equation \ref{eq:importance}. Thus, we can redefine the important samples: Important samples are samples that can represent most data of a class and samples that are difficult to distinguish by the model.

\subsubsection{Analysis of the confidence change }\quad

\mname can draw similar changing trends of confidence when training different models. As shown in Figure \ref{fig:confidence change}, the confidence first increases sharply, then drops and rises, eventually balances. It matches the overall trend of the Dunning-Kruger Effect curve and shows that we have realized the human confidence-imitated learning process. At the beginning of training, model weights are very random and the the classification results with arbitrarily invalid weights are almost the same. As the confidence is based on the uncertainty, it will lead to high self-confidence although the model is ignorant. Then the model weights are gradually formed and mature, dropping weights randomly will easily lead to different classification results, so the confidence is reduced. Finally, The weight is robust enough to avoid the result change caused by weight failure. As shown in Figure \ref{fig:confidence change}, low accuracy corresponds to blindly high confidence, fairly good accuracy corresponds to confused low confidence, and high accuracy corresponds to robust high confidence.

\section{Conclusion}

In this paper, we propose a new concept of Continuous Classification of Time Series (CCTS) to meet real needs. To solve its two problems about data arrangement and model training, we design a novel confidence-guided approach \mname. It can imitate the human behavior of object-confidence and self-confidence by the importance coefficient method and uncertainty evaluation. It arranges data iteration and reviews data according to importance-based object-confidence, and schedules training duration and training order according to uncertainty-based self-confidence. We test the method on four real-world datasets based on perspectives of classification accuracy, solving two difficulties, the difference among baselines, hyper-parameters, and analysis of data distributions and confidence change. The results show that our method is better than all baselines. It proves that the confidence-guided training strategy is an effective and self-adaptive indicator to guide the training and is more worthy of future research.

\section*{Acknowledgments}
This work was supported by the National Key Research and Development Program of China under Grant 2021YFE0205300, and the National Natural Science Foundation of China (No.62172018, No.62102008).

\bibliographystyle{unsrt}  
\bibliography{references}  

\clearpage

\appendix

\begin{center}
\Large \quad \\  \textbf{Appendix for \mname: \texttt{C}onfidence-guided Learning Process for \texttt{C}ontinuous \texttt{C}lassification of \texttt{T}ime \texttt{S}eries}\\
\quad \\

\end{center}

\normalsize
\section{Concept}

\subsection{Continuous Classification of Time Series (CCTS)}

\indent \indent Continuous Classification of Time Series (CCTS) aims to classify as accurately as possible with the evolution of time series at every time. 

Time series is one of the most common data forms, the popularity of time series classification has attracted increasing attention in many practical fields, such as healthcare and industry. In the real world, many applications require classification at every time. For example, in the Intensive Care Unit (ICU), critical patients' vital signs develop dynamically, the status perception and disease diagnosis are needed at any time. Timely diagnosis provides more opportunities to rescue lives. In response to the current demand, we propose a new task -- Continuous Classification of Time Series (CCTS). It aims to classify as accurately as possible at every time in time series.

\subsection{Concept Difference}

\subsubsection{One-shot Classification, Continuous Classification} \quad \vspace{+0.2cm}

The existing (early) classification of time series is the one-shot classification, where the classification is performed only once at the final or an early time. However, many real-world applications require continuous classification. For example, intensive care patients should be detected and diagnosed at all times to facilitate timely life-saving.

\begin{itemize}
   
   \item One-shot Classification.\textit{ A time series $X=\{x_{1},...x_{T}\}$ is labeled with classes $C$. OC classifies $X$ at a time $t, t\leq T$ with the minimum loss $\mathcal{L}(f(X_{1:t}),C)$.}
   
   \item  Continuous Classification. \textit{A time series is $X=\{x_{1},...x_{T}\}$. At time $t$, $x_{1:t}$ is labeled with class $c_{t}$. Continuous Classification classifies $x_{1:t}$ at every time $t=1,...,T$ with the minimum loss $\sum_{t=1}^{T}\mathcal{L}(f(x_{1:t}),c^{t})$.}

\end{itemize} 

\subsubsection{Classification of Time Series (CTS), Early Classification of Time Series(ECTS), Continuous Classification of Time Series (CCTS)} \quad \vspace{+0.2cm}

The popularity of time series classification has attracted increasing attention in many practical fields. The foundation is Classification of Time Series (CTS). It makes classification based on the full-length data. But in time-sensitive applications, Early Classification of Time Series (ECTS) is more critical, classifying at an early time. For example, early diagnosis helps for sepsis outcomes. But both of them classify only once and lean a single data distribution. In fact, CCTS is composed of multiple ECTS and the continuous classification is composed of multiple one-shot classification.

\begin{itemize}
    \item Classification of Time Series (CTS). \textit{A dataset of time series $\mathcal{D}=\{(X^{n},C^{n})\}_{n=1}^{N}$ has $N$ samples. Each time series $X^{n}$ is labeled with a class $C^{n}$, CTS classifies time series using the full-length data by model $f:f(X)\to C$.}
    
    \item Early Classification of Time Series (ECTS). \textit{A dataset of time series $\mathcal{D}=\{(X^{n},C^{n})\}_{n=1}^{N}$ has $N$ samples. Each time series $X^{n}={X_{t}}_{t=1}^{T}$ is labeled with a class $C^{n}$. ECTS classifies time series in an advanced time $t$ by model $f:f(\{x_{1},x_{2},...,x_{t}\})\to C$, where $t<T$.}
    
    \item Continuous Classification of Time Series (CCTS). \textit{A dataset of time series $\mathcal{D}=\{(X^{n},C^{n})\}_{n=1}^{N}$ has $N$ samples. Each time series $X^{n}={X_{t}}_{t=1}^{T}$ is labeled with a class $C^{n}$. CCTS classifies time series in every time $t$ by model  $f:f(\{x_{1},x_{2},...,x_{t}\})\to C$, where $t=1,...,T$.}
\end{itemize}

\subsubsection{Curriculum Learning (CL), Continual Learning (CL), Online Learning (OL), Continuous Classification of Time Series (CCTS)} 

\begin{itemize}
    \item Curriculum Learning (CL). \textit{A curriculum is a sequence of training criteria over $T$ training steps $\mathcal{C}=\{Q_{1},...,Q_{T}\}$. Each criterion $Q_{t}$ is a reweighting of the target training distribution $P(z)$, $Q_{t}(z)\propto W_{t}(z)P(z), \forall$ example $x\in$ training set $D$, such that the following three conditions are satisfied: (1) The entropy of distributions gradually increases, i.e., $H(Q_{t}<H(Q_{t+1}))$. (2) The weight for any example increases, i.e., $W_{t}(z)\leq W_{t+1}(z), \forall z\in D$. (3) $Q_{T}(z)=P(z)$.}
    
    \item Continual Learning (CL). \textit{A CL issue $\mathcal{T}=\{T^{1},T^{2},...,T^{N}\}$ has a sequence of $N$ tasks. Each task $T^{n}=(X^{n},C^{n})$ is represented by the training sample $X^{n}$ with classes $C^{n}$. CL learns a new task at every moment. The goal is to control the statistical risk of all seen tasks $\sum_{n=1}^{N}\mathbb{E}_{(X^{n},C^{n})}[\mathcal{L} (f_{n}((X^{n};\theta),C^{n})]$ with loss $\mathcal{L}$, network function $f_{n}$ and parameters $\theta$.}
    
    \item Online Learning (OL). \textit{A OL issue has a sequence of dataset $\mathcal{X}=\{X^{1},X^{2},...,X^{N}\}$ for one task $\mathcal{T}$. Each dataset $X^{t}$ has a distribution $D^{t}$. CL learns a new $D^{t}$ at every time $t$. The goal is to find the optimal solution of $\mathcal{T}$ after $N$ iterations by minimize the regret $\mathcal{R}:=\sum_{t=1}^{N}(f^{t}(X^{t})-\min f^{t}(X^{t}))$.}
    
    \item Continuous Classification of Time Series (CCTS). \textit{A dataset $\mathcal{S}$ contains $N$ time series. Each time series $X=\{x_{1},...x_{T}\}$ is labeled with a class $C\in \mathcal{C}$ at the final time $T$. As time series varies among time, it has a subsequence series with $M$ different distributions $\mathcal{D}=\{\mathcal{D}^{1},...,\mathcal{D}^{M}\}$, each $\mathcal{D}^{m}$ has subsequence $X_{1:t^{m}}$. CCTS learns every $\mathcal{D}^{m}$ and introduces a task sequence $\mathcal{M}=\{\mathcal{M}^{1},...,\mathcal{M}^{M}\}$ to minimize the additive risk $\sum_{m=1}^{M}\mathbb{E}_{\mathcal{M}^{m}}[\mathcal{L} (f^{m}(\mathcal{D}^{m};\theta),C)]$ with model $f$ and parameter $\theta$. $f^{m}$ is the model $f$ after being trained for $\mathcal{M}^{m}$. When the model is trained for $\mathcal{M}^{m}$, its performance on all observed data cannot degrade: $\frac{1}{m}\sum_{i=1}^{m}\mathcal{L}(f^{i},\mathcal{M}^{i}) \leq \frac{1}{m-1}\sum_{i=1}^{m-1} \mathcal{L}(f^{i},\mathcal{M}^{i})$.}
\end{itemize}

\subsubsection{Data Stream Classification, Multi-step Prediction, Anytime Classification}

\begin{itemize}
    \item Data stream (big data, focus on the current) and time series (a whole data, focus on the overall) are two different data form. Data stream classification aims to the rapid feature extraction and model optimization for the arriving data, while our CCTS can classify new time series anytime without training again. 
    
    \item Multi-step prediction predicts values in multiple future time steps, similar to the early classification of time series (ECTS). But our CCTS aims to predict the final label, like patient outcome, at every time. CCTS has to solve the problem of catastrophic forgetting, but ECTS doesn't. 

    \item ‘Anytime’ is another expression of ‘continuous’.
\end{itemize}

\section{Experiments}

\subsection{Datasets} 

\begin{itemize}

    \item One data corresponds to one label, e.g. the label `sepsis' for a Sepsis data and `mortality' for a COVID-19 data. We copy the final label of the time series to its each time point.
        \begin{itemize}
        \item Sepsis dataset \cite{DBLP:conf/cinc/ReynaJSJSWSNC19} has 30,336 records with 2,359 diagnosed sepsis. Early diagnose is critical to improve sepsis outcome \cite{seymour2017time}. In this dataset, the time series are the changes of 40 related patient features, the label at each time is sepsis or non-sepsis. 
        \item COVID-19 dataset \cite{COVID-19} has 6,877 blood samples of 485 COVID-19 patients from Tongji Hospital, Wuhan, China. Mortality prediction helps for treatment and rational resource allocation \cite{DBLP:journals/BMC/sun}. In this dataset, the time series are the changes of blood samples, the label at each time is mortality or survival. 
        \item UCR-EQ dataset has 471 earthquake records from UCR time series database archive. It is the univariate time series of seismic feature value. Natural disaster early warning, like earthquake warning, helps to reduce casualties and property losses.
        \item USHCNrain dataset \cite{USHCN} has the daily meteorological data of 48 states in U.S. from 1887 to 2014. It is the multivariate time series of 5 weather features. Rainfall warning is not only the demand of daily life, but also can help prevent natural disasters.
        \end{itemize}
        Not that for each time series in the above four datasets, every time point is tagged with a class label, which is the same as its outcome label, such as `mortality', `sepsis', `earthquake' and `rain'.

    \item Different time points of a time series have different labels.
        \begin{itemize} 
        \item MIMIC-III dataset \cite{johnson2016mimic} has 19,993 admission records of 7,537 patients. We focus on 10 diagnoses (ICD-9): HIV (042), Brain Cancer (191), Diabetes(249), Hypertension (401), Heart Failure (428), Pneumonia (480-486), Gastric Ulcer (531), Hepatopathy (571), Nephropathy (580-589), SIRS (995.9). The time series are vital signs, and labels at each time are some diagnoses.
        \item USHCN dataset \cite{USHCN} has U.S. daily meteorological data from 1887 to 2014. We focus on 4 weather conditions in New York: sunny, overcast, rainfall, snowfall. The time series are records of 4 neighboring states, labels at each time are weather after a weak.
        \end{itemize}
\end{itemize}

\subsection{Baselines}

\noindent \textbf{Baselines}
\begin{itemize} 
    \item Early classification based methods:
    
        \begin{itemize}
        \item LSTM \cite{choi2017using} uses a base model trained by time series at every time stage.
        \item SR \cite{DBLP:journals/tnn/MoriMDL18} applies multiple base models trained by the full-length time series and uses the fusion result.
        \item ECEC \cite{lv2019effective} has a set of base classifiers trained by time series in different time stages.
        \end{itemize}
        
    \item Curriculum learning based methods:

        \begin{itemize} 
        \item DIF \cite{DBLP:conf/icml/HacohenW19}. It arranges curriculum (data learning order/task order) by loss/difficulty getting from the teacher model.
        \item UNCERT \cite{2021A}. It arranges curriculum (data learning order/task order) by sentence uncertainty and model uncertainty.
        \end{itemize}
        
    \item  Continual learning based methods:
        \begin{itemize}
        \item EWC \cite{DBLP:journals/corr/KirkpatrickPRVD16}. It is a regularization-based method, training a model to remember the old tasks by constraining important parameters to stay close to their old values.
        \item GEM \cite{DBLP:conf/nips/Lopez-PazR17} trains a base model to remember the old tasks by finding the new gradients which are at acute angles to the old gradients.
        \item CLEAR. It is a replay-based method, using the reservoir sampling to limit the number of stored samples to a fixed budget assuming an i.i.d. data stream.
        \item CLOPS \cite{2021A} trains a base model by replaying old tasks with importance-guided buffer storage and uncertainty-based buffer acquisition.
        \end{itemize}
        
     \item Online learning based methods:
     
        \begin{itemize} 
        \item OSFW \cite{DBLP:conf/icml/ChenHHK18} trains a base model by stochastic gradient estimator to achieves the stochastic regret bound.
        \item ORGFW \cite{DBLP:conf/aaai/XieSZWQ20} trains the model by a recursive gradient estimator to achieve an optimal regret bound.
        \end{itemize}
        
    \item Confidence based methods:

        \begin{itemize}
        \item DROPOUT \cite{DBLP:conf/icml/GalG16}. It uses the Monte Carlo Dropout method to approximate Bayesian inference and gets the model uncertainty.
        \item TCP \cite{DBLP:conf/nips/CorbiereTBCP19}. It designs a true class probability to estimate the confidence of the model prediction.
        \item STL \cite{DBLP:journals/corr/abs-2105-11545}. It gives the uncertainty about the signal temporal logic, which is more suitable for time series.
        \end{itemize}
\end{itemize}

\clearpage

\begin{table*}[h]
\caption{Classification Accuracy (AUC-ROC$\uparrow$) of Baselines for 4 Real-world Datasets at 10 Time Steps.\newline
\small{$^{1}$The dataset has one task. Use online learning methods, OSFW and ORGFW, as baselines. 
$^{2}$The dataset has multiple tasks. Use continual learning methods, GEM and CLOPS, as baselines.
$^{3}$All methods use LSTM as the base model for fairness.
$^{4}$k\% means the current classification time is k\% of the total time of the full-length time series.
$^{5}$The value is the average accuracy from 5-fold cross-validation with mean±std. }
} \label{tb:appendix1}
\centerline{
\small
\setlength{\tabcolsep}{0.3mm}{
\begin{tabular}{llllllllllll}
\toprule[0.8pt]
 Dataset &\tiny{\diagbox{Method$^{3}$}{Time$^{4}$}} &10\%    &20\%   &30\%    &40\%    &50\%    &60\%   &70\%    &80\%    &90\%    &100\%  \\
\midrule[0.8pt]
\multirow{6}*{Sepsis$^{1}$} 
& LSTM &0.576±0.060$^{5}$    &0.629±0.031    &0.715±0.060    &0.736±0.061    &0.745±0.053    &0.748±0.044    &0.773±0.035    &0.795±0.025    &0.813±0.022    &0.827±0.013      \\
& SR &0.626±0.036    &0.659±0.031    &0.738±0.0100    &0.761±0.020    &0.803±0.0010    &0.807±0.030    &0.815±0.010    &0.835±0.010    &0.835±0.020    &0.850±0.020     \\
& ECEC &0.623±0.020    &0.669±0.010    &0.731±0.010    &0.763±0.010    &0.811±0.010    &0.815±0.010    &0.820±0.010    &0.823±0.010    &0.839±0.010    &0.851±0.010     \\
& OSFW &0.670±0.010    &0.712±0.020     &0.735±0.010    &0.765±0.020    &0.821±0.030     &0.825±0.020     &0.839±0.010    &0.850±0.010    & 0.851±0.010     &0.861±0.010      \\
& ORGFW &0.670±0.020    &0.714±0.020    &0.732±0.020    &0.766±0.030    &0.820±0.020   &0.831±0.020    &0.832±0.030    &0.843±0.010    &0.850±0.010    &0.862±0.010    \\
& DIF  &0.666±0.025 &0.710±0.021  &0.732±0.024   &0.755±0.027    &0.825±0.025   &0.832±0.027   &0.839±0.028      &0.844±0.012 &0.847±0.010      &0.848±0.016    \\
& UNCERT  &0.665±0.026 &0.705±0.019  &0.733±0.025  &0.759±0.028     &0.824±0.029   &0.831±0.027   &0.838±0.026   &0.846±0.015    &0.850±0.017    &0.857±0.018    \\
& DROPOUT  &0.660±0.021 &0.766±0.015  &0.720±0.014    &0.748±0.021    &0.820±0.020     &0.825±0.022  &0.832±0.021  &0.835±0.018    &0.840±0.015   &0.850±0.011    \\
& TCP  &0.662±0.021 &0.705±0.016  &0.722±0.020   &0.758±0.025    &0.826±0.027    &0.827±0.027  &0.839±0.025      &0.840±0.017 &0.843±0.011      &0.862±0.016    \\
& STL   &0.660±0.023 &0.709±0.016  &0.727±0.021    &0.760±0.024   &0.824±0.026     &0.833±0.028      &0.836±0.015 &0.840±0.016  &0.845±0.016    &0.855±0.017    \\
&\textbf{\mname} &\textbf{0.671±0.025}   &\textbf{0.715±0.024} &\textbf{0.734±0.021}   &\textbf{0.768±0.026}   &\textbf{0.831±0.023}       &\textbf{0.840±0.024}    &\textbf{0.840±0.018}  &\textbf{0.851±0.012}  &\textbf{0.853±0.012}    & \textbf{0.872±0.012}  \\

\midrule[0.4pt]
\multirow{6}*{COVID-19$^{1}$} 
& LSTM &0.605±0.04    &0.701±0.03     &0.793±0.02    &0.833±0.01    &0.844±0.01     &0.888±0.01    &0.918±0.03    &0.925±0.01    &0.939±0.00     &0.944±0.01       \\
& SR &0.636±0.01    &0.730±0.02     &0.810±0.01    &0.867±0.01    &0.901±0.01     &0.900±0.01     &0.935±0.01    &0.946±0.00    &0.952±0.01     &0.962±0.00   \\
& ECEC &0.639±0.01    &0.732±0.02     &0.829±0.01    &0.870±0.01    &0.901±0.02     &0.904±0.01     &0.937±0.00    &0.948±0.01    &0.952±0.00     &0.955±0.01   \\
& OSFW &0.693±0.02    &0.765±0.01     &0.865±0.01    &0.887±0.02    &0.915±0.01     &0.913±0.01     &0.941±0.00    &0.955±0.01    &0.960±0.01     &0.964±0.00      \\
& ORGFW &0.709±0.01    &0.775±0.01     &0.849±0.01    &0.878±0.01   &0.916±0.02     &0.912±0.01     &0.945±0.01    &0.957±0.00    &0.961±0.00     &0.965±0.00       \\
& DIF    &0.785±0.019  &0.830±0.021  &0.862±0.017  &0.879±0.016  &0.915±0.014  &0.926±0.014       &0.936±0.010 &0.941±0.007 &0.947±0.006   &0.952±0.006       \\
& UNCERT   &0.775±0.013  &0.841±0.016  &0.871±0.015  &0.900±0.013  &0.915±0.013 &0.925±0.015     &0.935±0.009 &0.940±0.007  &0.950±0.005   &0.954±0.006       \\
& DROPOUT    &0.740±0.017  &0.831±0.020 &0.860±0.013   &0.885±0.012  &0.912±0.011    &0.924±0.018   &0.932±0.011   &0.939±0.007  &0.943±0.004  &0.945±0.005      \\
& TCP    &0.786±0.012   &0.832±0.019  &0.872±0.018  &0.895±0.016   &0.915±0.014 &0.927±0.011       &0.941±0.010 &0.942±0.007   &0.948±0.004 &0.947±0.008       \\
& STL   &0.770±0.012   &0.833±0.017  &0.870±0.011 &0.895±0.014  &0.916±0.013 &0.925±0.013     &0.941±0.012 &0.944±0.011   &0.948±0.007 &0.950±0.008       \\
&\textbf{\mname}  &\textbf{0.790±0.021}  &\textbf{0.843±0.19}   &\textbf{0.877±0.020}  &\textbf{0.901±0.015}  &\textbf{0.919±0.015}   &\textbf{0.927±0.012}   &\textbf{0.945±0.011}   &\textbf{0.960±0.011}   &\textbf{0.967±0.010}  &\textbf{0.969±0.010} \\

\midrule[0.7pt]
\multirow{6}*{MIMIC-III$^{2}$}
&\ LSTM &0.574±0.04    &0.595±0.04      &0.635±0.01     &0.661±0.02    &0.706±0.02      &0.724±0.02      &0.770±0.01    &0.785±0.01     &0.793±0.01      &0.817±0.01    \\
 &\ SR &0.628±0.02     &0.681±0.01      &0.695±0.01     &0.692±0.01     &0.732±0.01      &0.749±0.01      &0.785±0.01    &0.805±0.01     &0.820±0.01      &0.825±0.01    \\
&\ ECEC &0.651±0.02     &0.702±0.01      &0.699±0.01     &0.703±0.01     &0.733±0.01      &0.756±0.01      &0.792±0.01    &0.808±0.01     &0.825±0.01      &0.838±0.01    \\
&\ GEM &0.685±0.01     &0.714±0.02      &0.732±0.01     &0.747±0.01     &0.759±0.01      &0.772±0.01      &0.805±0.01    &0.816±0.01     &0.826±0.01      &0.827±0.01    \\
&\ CLOPS &0.691±0.01     &0.722±0.01      &0.743±0.01     &0.752±0.01     &0.764±0.01      & 0.770±0.01     &0.808±0.02    &0.815±0.01     &0.825±0.01      &0.831±0.01      \\
&\textbf{\mname} &\textbf{0.714±0.02 } &\textbf{0.731±0.02}    &\textbf{0.757±0.01 }    &\textbf{0.763±0.01 }    &\textbf{0.780±0.01 }     &\textbf{0.788±0.01 }    &\textbf{0.810±0.02}   &\textbf{0.820±0.01}    &\textbf{0.832±0.01 }   &\textbf{0.844±0.01 }\\

\midrule[0.4pt]
\multirow{6}*{USHCN$^{2}$}
&\ LSTM &0.618±0.03    &0.621±0.03     &0.674±0.03    &0.694±0.03    &0.724±0.04     &0.750±0.02      &0.775±0.02    &0.793±0.03    &0.856±0.02      &0.861±0.01    \\
&\ SR &0.674±0.03    &0.679±0.01   &0.706±0.02    &0.722±0.02     &0.752±0.01      &0.767±0.01      &0.805±0.01    &0.831±0.01     &0.866±0.01      &0.885±0.01    \\
&\ ECEC &0.688±0.03    &0.701±0.02     &0.721±0.02    &0.730±0.01    &0.769±0.01      &0.781±0.01      &0.806±0.02    &0.839±0.01     &0.870±0.01      &0.889±0.01    \\
&\ GEM &0.709±0.02    &0.741±0.01     &0.763±0.01    &0.762±0.02    &0.780±0.01     &0.809±0.02      &0.825±0.01    &0.843±0.01     &0.877±0.01      &0.889±0.01        \\
&\ CLOPS &0.721±0.02    &0.740±0.02     &0.763±0.01    &0.771±0.02    &0.772±0.02     &0.805±0.01     &0.830±0.02    &0.847±0.01     &0.875±0.01      &0.897±0.01      \\
&\textbf{ \mname} &\textbf{0.732±0.01} &\textbf{0.750±0.01}    &\textbf{0.767±0.01}    &\textbf{0.780±0.02}   &\textbf{0.783±0.01}     &\textbf{0.811±0.01}     &\textbf{0.848±0.02}    &\textbf{0.860±0.01 }    &\textbf{0.884±0.01}   &\textbf{0.902±0.01}\\
\bottomrule[0.8pt]
\end{tabular}
}}
\end{table*}

\begin{table*}[!ht]
\caption{COVID-19 Classification Accuracy with Non-uniform Training Sets and Validation Sets.
\footnotesize {$\downarrow$ means the accuracy is greatly reduced}} \label{tb:COVID-19 overfitting}
\centering
\small
\setlength{\tabcolsep}{2mm}{
\begin{tabular}{llllllll}
\toprule[0.8pt]
Subset   &SR  &ECEC    &CLEAR    &CLOPS    &DIF &UNCERT    &\textbf{\mname}   \\
\midrule[0.8pt]
Male   &0.968±0.014   &0.969±0.016   &0.965±0.012 &0.965±0.004 &0.978±0.009 &0.978±0.014   &0.971±0.010\\
Female    &0.915±0.004$\downarrow$   &0.917±0.015$\downarrow$   &0.919±0.018 &0.928±0.003 &0.919±0.008$\downarrow$ &0.941±0.009 &0.947±0.002\\
\midrule[0.4pt]
Age 30-   &0.965±0.014   &0.967±0.015   &0.967±0.013 &0.964±0.009 &0.977±0.008 &0.979±0.012   &0.972±0.010\\
Age 30+  &0.941±0.007   &0.943±0.018   &0.931±0.008$\downarrow$ &0.923±0.040$\downarrow$ &0.902±0.006$\downarrow$ &0.914±0.007$\downarrow$   &0.945±0.006\\
\midrule[0.4pt]
Test    &0.964±0.013   &0.968±0.015   &0.966±0.012 &0.962±0.006 &0.979±0.009 &0.978±0.010   &0.970±0.007\\
Valid.   &0.962±0.006   &0.963±0.014   &0.954±0.003 &0.953±0.005 &0.952±0.009$\downarrow$ &0.954±0.004$\downarrow$   &0.967±0.006\\
\bottomrule[0.8pt]
\end{tabular}
}
\end{table*}

\begin{figure*}[h]
\centering
\caption{Different Concepts}
\includegraphics[width=\linewidth]{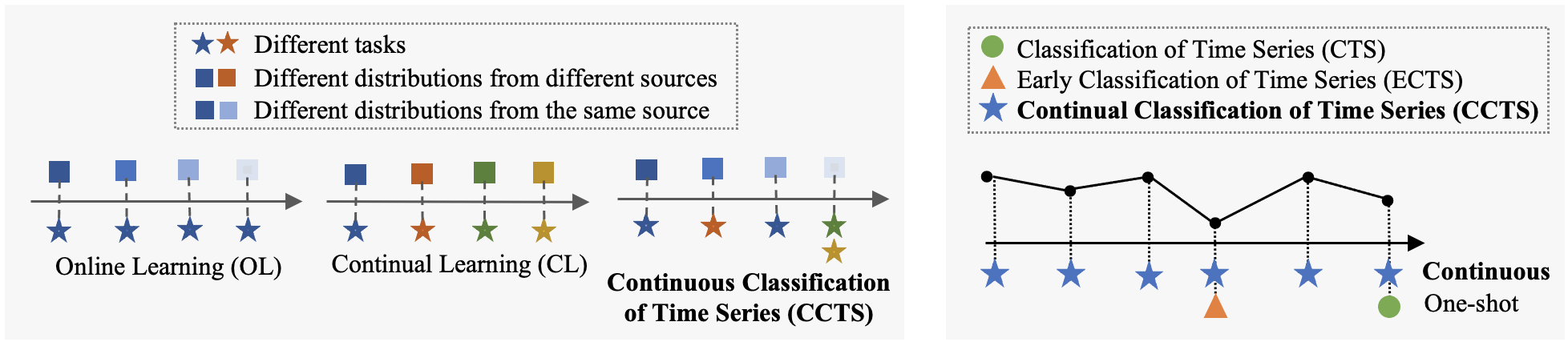}
\label{fig:concept}
\end{figure*}

\begin{figure*}[h]
\centering
\caption{Multi-distribution in MIMIC-III Dataset}
\includegraphics[width=0.95\linewidth]{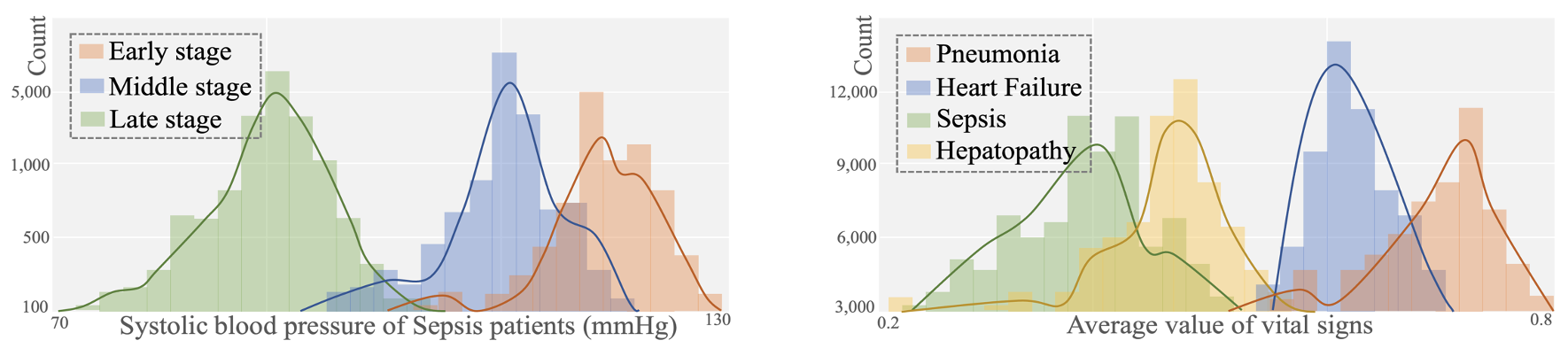}
\end{figure*}

\begin{figure*}[h]
\centering
\caption{The Important Samples in SEPSIS Distribution Buffers}
\includegraphics[width=\linewidth]{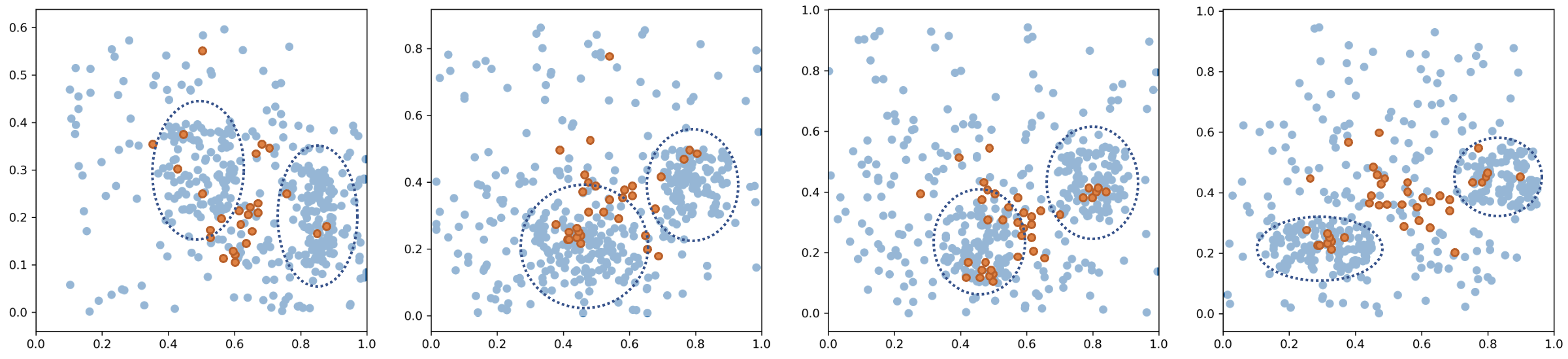}
\end{figure*}

\begin{figure}[h]
\caption{Task Similarity \newline
\small{In MIMIC-III dataset, the diagnoses with ICD-9 order are 1:HIV, 2:Brain Cancer, 3:Diabetes, 4:Hypertension, 5:Heart Failure, 6:Pneumonia, 7:Gastric Ulcer, 8:Hepatopathy, 9:Nephropathy, 10:SIRS. The new similarity order is 1, 10, 2, 4, 5, 8, 3, 9, 7, 6.}}
\centering
\includegraphics[width=0.5\linewidth]{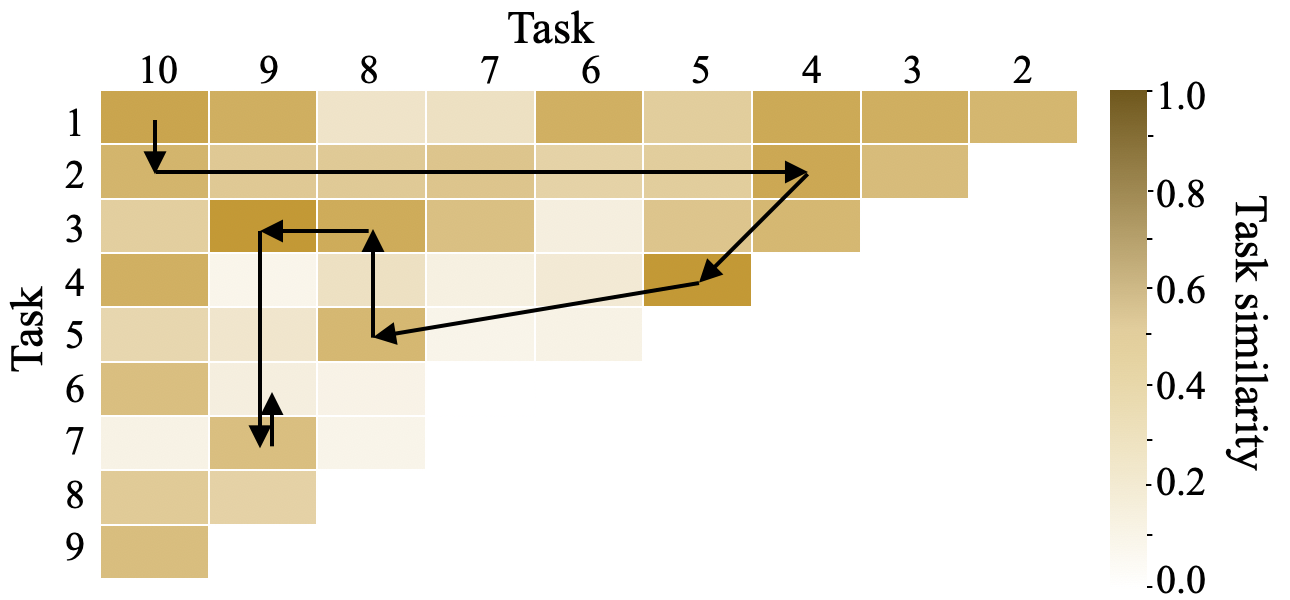}
\end{figure}

\end{document}